\RequirePackage[svgnames,table]{xcolor}
\documentclass[11pt, letterpaper, logo]{merged_style}

\usepackage[comma,authoryear,compress]{natbib}

\usepackage{amsmath,amsfonts,bm}

\def\eqref#1{equation~\ref{#1}}

\def\1{\bm{1}}

\DeclareMathAlphabet{\mathsfit}{\encodingdefault}{\sfdefault}{m}{sl}
\SetMathAlphabet{\mathsfit}{bold}{\encodingdefault}{\sfdefault}{bx}{n}

\usepackage{algorithm}
\usepackage{algorithmic}
\usepackage{subcaption}
\usepackage{multirow}
\usepackage{placeins}
\usepackage{soul}
\usepackage{etoolbox}
\AtBeginEnvironment{algorithmic}{\small}
\AtBeginEnvironment{proof}{\sloppy}
\usetikzlibrary{arrows.meta,calc,positioning,shapes.geometric}

\definecolor{revisionblue}{RGB}{218,235,255}
\sethlcolor{revisionblue}

\newcommand{\algorithmtoprule}{
    \par\noindent\rule{\linewidth}{1.5pt}\par
    \vspace{-0.5em}
}

\newcommand{\algorithmmidrule}{
    \vspace{-0.5em}
    \par\noindent\rule{\linewidth}{0.5pt}\par
}

\newcommand{\algorithmbottomrule}{
    \vspace{-0.5em}
    \par\noindent\rule{\linewidth}{1pt}\par
    \vspace{0.5em}
}

\title{What to Practice Next? Latent Profile-Guided Task Selection for Skill-Evolving LLM Agents}
\runningtitle{What to Practice Next? Latent Profile-Guided Task Selection for Skill-Evolving LLM Agents}

\author{\begin{minipage}{\textwidth}
\centering\fontsize{13}{14}\selectfont\bfseries
Qijin Zhang\textsuperscript{\normalfont *,1}, \quad
Hanzhen Wang\textsuperscript{\normalfont *,1}, \quad
Yan Huang\textsuperscript{\normalfont 2,3}\par
Zhongxiang Dai\textsuperscript{\normalfont 4}, \quad
Shuang Qiu\textsuperscript{\normalfont \textdagger,1}
\par\vspace{5pt}
{\fontsize{11}{13}\selectfont\normalfont
\textsuperscript{\normalfont 1}City University of Hong Kong\par
\textsuperscript{\normalfont 2}Institute of Automation, Chinese Academy of Sciences\par
\textsuperscript{\normalfont 3}Five Ages\par
\textsuperscript{\normalfont 4}The Chinese University of Hong Kong, Shenzhen\par\vspace{5pt}
\texttt{qijizhang5-c@my.cityu.edu.hk} \quad \texttt{hanzhwang9-c@my.cityu.edu.hk}\par
\texttt{yhuang@nlpr.ia.ac.cn} \quad \texttt{daizhongxiang@cuhk.edu.cn}\par
\texttt{shuanqiu@cityu.edu.hk}}
\end{minipage}}
\bannerlogos{%
\hfill
  \includegraphics[height=11mm]{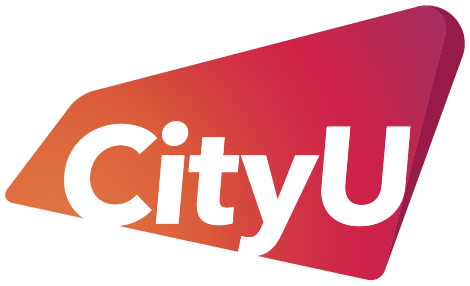}\hspace{10pt}%
  \includegraphics[height=11mm]{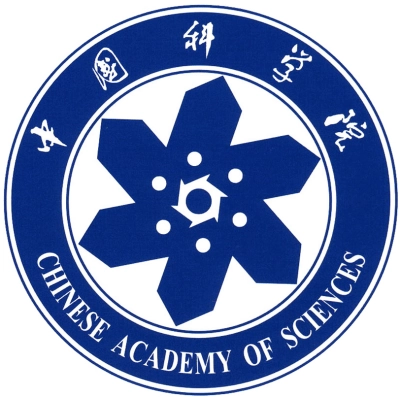}\hspace{10pt}%
  \includegraphics[height=11mm]{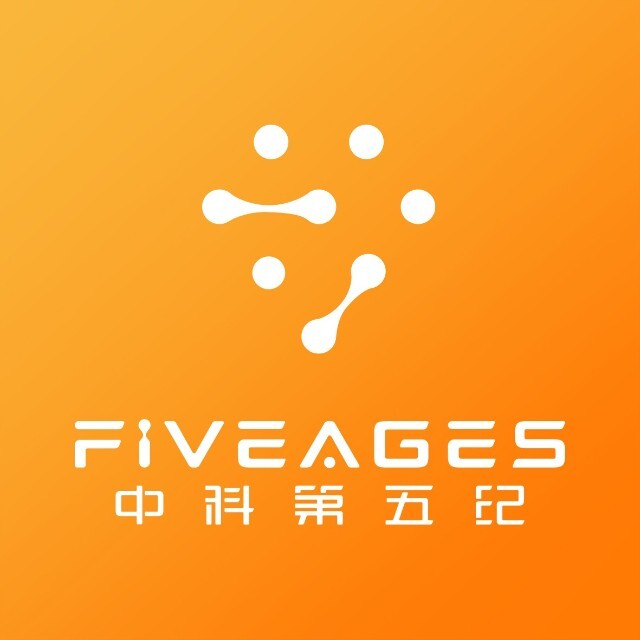}\hspace{10pt}%
  \includegraphics[height=12mm]{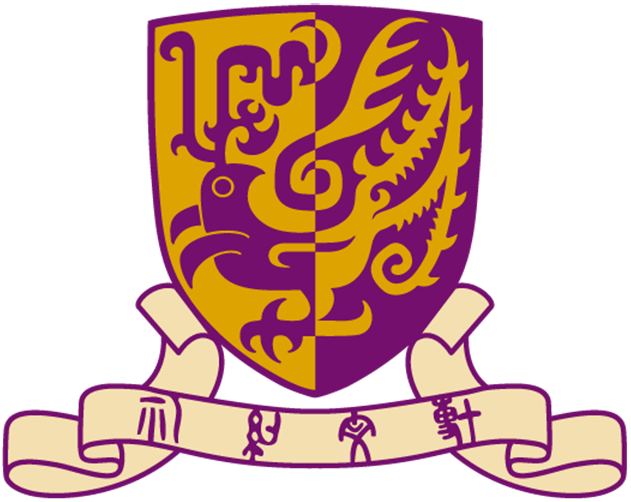}%
\vspace{-1.1cm}
}

\begin{abstract}
Task selection shapes the experiences available for skill evolution in LLM agents, yet fixed practice schedules overlook differences in skill requirements and evolving task value. Most existing methods, however, rely on fixed schedules or scalar measures of progress, leaving nonstationary skill demands unaddressed. We propose SPG-Select, a framework for allocating limited practice budget using a multidimensional profile inferred from binary outcomes via MIRT. The framework predicts surrogate progress, defined as the difference between consecutive profile estimates, and prioritizes progress in dimensions with larger estimated proficiency gaps. Our SPG-Select-R policy combines these predictions with uncertainty-based exploration and periodic restarting to accommodate nonstationarity. We evaluate SPG-Select on ALFWorld and WebShop with three skill-evolution backends. In the reported runs, SPG-Select-R achieves higher mean held-out terminal success than uniform sampling in all six backend-benchmark combinations, with gains of 1.49-13.18 percentage points on ALFWorld and 0.93-3.13 points on WebShop. Under a predictable linear surrogate model and boundedness assumptions, we establish a high-probability dynamic pseudo-regret bound for surrogate task allocation in terms of parameter path variation.
\end{abstract}

\begin{document}
\maketitle
\begingroup
\renewcommand{\thefootnote}{}\NoHyper
\footnotetext{\begin{tabular}[t]{@{}l@{}}
\textsuperscript{*}Equal contribution.\\
\textsuperscript{\textdagger}Corresponding author.
\end{tabular}}\endNoHyper
\endgroup

\addtocontents{toc}{\protect\setcounter{tocdepth}{-1}}

\section{Introduction}
\label{sec:introduction}

Large language model (LLM) agents can improve through repeated interaction, reflection, and reuse of accumulated experience \citep{Yao2022a,Shinn2023,Zhao2024,Wang2023}. Recent work further frames self-evolution as a lifelong process coupling experience exploration, persistent memory, and reusable skill learning \citep{Cai2025,Wu2025,Xia2026}. In this work, skill evolution refers to extracting reusable knowledge from trajectories for subsequent decisions. Since the quality of skill evolution depends on which experiences are acquired as well as how they are summarized and retrieved, practice-task selection is a central design decision.

With a limited interaction budget, maximizing immediate success can overallocate practice to already-mastered tasks, whereas more challenging tasks may better develop weak abilities even when they initially fail. Different tasks can also affect skill dimensions unevenly. This motivates ranking tasks by expected surrogate progress relative to the current profile estimate rather than by immediate success or task difficulty alone.

The policy observes only binary outcomes, which reveal neither the contributing latent dimensions nor the true skill change after a backend update. We therefore infer a multidimensional profile as an estimate of the latent skill state and model temporal differences between consecutive profile estimates online. Because accumulated knowledge can change subsequent outcomes and hence the distribution of these temporal differences, older observations can become stale. Task selection is consequently an allocation problem with indirect, nonstationary feedback rather than a static ranking problem or a conventional contextual bandit problem with directly observed scalar rewards.

Curriculum learning has shown that task order can affect optimization and generalization \citep{Bengio2009}, while automatic curricula adapt task distributions to evolving capabilities \citep{Portelas2020}. Self-Evolving Curriculum uses a nonstationary bandit over problem categories during LLM reinforcement-learning fine-tuning \citep{Chen2025}, while Experience-Driven Lifelong Learning couples experience exploration, skill abstraction, and memory evolution \citep{Cai2025}. We instead study finite-budget task selection from a fixed heterogeneous task pool, using a profile inferred from binary outcomes as an estimate of the latent skill state.

We treat each interaction as both an assessment that updates the profile estimate and an intervention that may affect subsequent behavior through backend updates. Online multidimensional item response theory (MIRT) produces this latent profile estimate rather than a directly observed skill vector, while a contextual bandit selects practice tasks using predicted dimension-wise surrogate progress, i.e., temporal differences between consecutive profile estimates, together with uncertainty. Section~\ref{sec:method} explains why MIRT is structurally suited to this role. Its Bernoulli observation model matches the available binary feedback, it represents task difficulty separately from profile-conditioned proficiency, and its vector-valued profile enables the dimension-wise gap weighting used by the policy. The policy thereby targets profile-conditioned surrogate progress rather than task difficulty or success probability alone. We call this framework Skill Profile-Guided Task Selection (SPG-Select). We model task selection as a nonstationary contextual bandit, allowing the conditional mean of surrogate progress to vary across rounds. To limit stale observations, we instantiate SPG-Select with a sliding-window upper confidence bound (UCB) \citep{Cheung2019} as SPG-Select-W and with periodic restarting \citep{Zhao2020} as SPG-Select-R. We examine whether this decision layer applies across a skill bank, an insight bank, and shared reflection, answering the question: \emph{can profile-guided task selection improve the use of a limited
practice budget?}

To address this question, we make four contributions. First, we formulate finite-budget task selection from a fixed pool as a nonstationary contextual bandit. Temporal differences between MIRT profile estimates provide vector-valued surrogate feedback, which does not directly measure true skill change. A profile-gap-weighted scalar reward emphasizes surrogate progress in dimensions with lower estimated proficiency. Second, we develop SPG-Select, which combines MIRT profile estimation, a task-profile encoder, vector-valued surrogate-progress prediction, and gap-weighted UCB selection under nonstationarity. Third, we evaluate the selector with SkillRL, Experiential Learning (ExpeL), and Shared Reflexion on ALFWorld and WebShop. In the reported runs, SPG-Select-R achieves higher mean held-out success than Uniform in all six backend-benchmark combinations. Component ablations assess the uncertainty bonus and profile-gap weighting. Fourth, under a predictable time-varying linear surrogate model and boundedness assumptions, we establish a high-probability dynamic pseudo-regret bound for SPG-Select-R in terms of parameter path variation and characterize the oracle restart-period tradeoff. The guarantee concerns surrogate task allocation.

\noindent\textbf{Related work.}
Our study connects experience-based skill evolution in LLM agents, including linguistic self-feedback, transferable insight extraction, and reusable skill learning \citep{Shinn2023,Zhao2024,Xia2026}, with multidimensional ability estimation through MIRT and adaptive assessment \citep{Reckase2006,Sharpnack2024}. Experience-based methods provide mechanisms for evolving knowledge from selected practice trajectories, while MIRT provides a latent-state estimate, denoted by the profile, for interpreting binary outcomes. We use this profile for task allocation by drawing on contextual and nonstationary bandits, including linear UCB, sliding-window estimation, and periodic restarting to address parameter drift \citep{AbbasiYadkori2011,Cheung2019,Zhao2020}. Appendix~\ref{app:related-work} provides a fuller account of related work.

\section{Problem Setup}
\label{sec:problem}

Consider an LLM agent interacting with a fixed task pool used for evolution \(\mathcal{T}=\{\tau^{(1)},\tau^{(2)},\dots,\tau^{(M)}\}\) consisting of \(M\in\mathbb{N}_+\) tasks over \(T\in\mathbb{N}_+\) successive bandit rounds indexed by \(t\in[T]=\{1,\ldots,T\}\). All tasks in \(\mathcal{T}\) remain available at every round. Immediately before the decision at round \(t\), the agent has a true latent skill state \(\boldsymbol{s}_t^\star\in[0,1]^K\). This state is assumed to exist, represents the agent's proficiency in \(K\in\mathbb{N}_+\) skill dimensions, and cannot be directly observed. We use the term profile exclusively for the MIRT estimate of this latent state. Specifically, the pre-decision profile is the predictive mean \(\widehat{\boldsymbol{s}}_t\in[0,1]^K\) inferred from all observations available before round \(t\). The agent selects a task \(\tau_t\in\mathcal{T}\), executes it, and observes a binary environmental outcome \(y_t\in\{0,1\}\), where $1$ denotes success and $0$ denotes failure. Following this interaction and the ensuing skill-evolution update, the unobserved latent state becomes \(\boldsymbol{s}_{t+1}^\star\), giving the true skill change \(\boldsymbol{\Delta}_t^\star=\boldsymbol{s}_{t+1}^\star-\boldsymbol{s}_t^\star\). Separately, the online MIRT filter uses \(y_t\) to produce the next pre-decision profile estimate \(\widehat{\boldsymbol{s}}_{t+1}\). We define the observable bandit feedback, surrogate progress, as the temporal difference between consecutive profile estimates,
\[
    \widehat{\boldsymbol{\delta}}_t
    :=\widehat{\boldsymbol{s}}_{t+1}-\widehat{\boldsymbol{s}}_t.
\]
Thus, \(\widehat{\boldsymbol{s}}_t\) denotes a profile estimate and \(\widehat{\boldsymbol{\delta}}_t\) denotes the corresponding temporal difference. The latter is an outcome-based proxy that reflects both belief revision and the state-transition model. It does not directly measure \(\boldsymbol{\Delta}_t^\star\) or the effect of the backend update. We use this surrogate to define the practice-allocation objective below.

Each task \(\tau\in\mathcal{T}\) is associated with a fixed task embedding \(\boldsymbol{e}_\tau\in\mathbb{R}^{d_e}\), where \(d_e\in\mathbb{N}_{+}\) denotes the embedding dimension. We obtain this embedding by applying a sentence encoder to the task's natural-language description. Before choosing a task at round \(t\), the agent therefore has access to the current profile \(\widehat{\boldsymbol{s}}_t\) and the embedding \(\boldsymbol{e}_\tau\) of every candidate task. We formulate adaptive task selection as a nonstationary contextual bandit problem, with context \((\widehat{\boldsymbol{s}}_t,\boldsymbol{e}_\tau)\) for candidate \(\tau\). Skill evolution can change the underlying latent state, while successive outcomes update the profile estimate and may alter the expected surrogate progress associated with practicing a task. The task pool and embeddings remain fixed. To emphasize surrogate progress in dimensions with larger estimated proficiency gaps, we define the gap-weight vector and the corresponding scalar reward as
\begin{equation}
    \widehat r_t = \widehat{\boldsymbol{g}}_t^\top\widehat{\boldsymbol{\delta}}_t,
    \quad
    \text{where } \widehat{\boldsymbol{g}}_t = \operatorname{softmax}\!\left( \frac{\mathbf{1} - \widehat{\boldsymbol{s}}_t}{\gamma_g} \right),
    \label{eq:gap-weights}
\end{equation}
where \(\mathbf{1}\in\mathbb{R}^K\) is the all-ones vector, \(\gamma_g>0\) is a temperature parameter, and \(\operatorname{softmax}\) exponentiates each coordinate and normalizes the vector to sum to one. Among any two dimensions, the one with lower estimated proficiency receives a larger weight. Thus, equal surrogate progress receives greater weight on a dimension with a larger estimated gap. A smaller \(\gamma_g\) concentrates the weights more strongly on the largest gaps, whereas a larger value distributes them more evenly. Since the common unit target cancels in the normalization, the weights depend on relative proficiency differences. The gap vector is computed before action selection. Its entries are nonnegative and sum to one, therefore \(\lVert\widehat{\boldsymbol{g}}_t\rVert_2\leq\lVert\widehat{\boldsymbol{g}}_t\rVert_1=1\). Also, \(\widehat{\boldsymbol{\delta}}_t\in[-1,1]^K\) implies \(\widehat r_t\in[-1,1]\), independently of \(K\).

We consider deterministic policies based on the current maintained state, comprising the profile estimator and online task-selection statistics described in Section~\ref{sec:method}. Past observations enter decisions through these recursively updated quantities. Let \(\Pi\) denote the policy class and write \(\pi=(\pi_t)_{t=1}^{T}\), where \(\pi_t\) is the round-specific decision map with its current internal state left implicit. Its explicit inputs are the current estimated profile and the fixed candidate-task embeddings. The objective is to maximize the expected cumulative reward over a bandit horizon of \(T\) rounds,
\[
    \sup_{\pi\in\Pi}\ \mathbb{E}_{\pi}\!\left[\sum_{t=1}^{T} \widehat r_t\right],
    \qquad
    \tau_t=\pi_t\!\left(\widehat{\boldsymbol{s}}_t,\{\boldsymbol{e}_\tau\}_{\tau\in\mathcal{T}}\right).
\]
The expectation is over the environmental and estimation randomness along the trajectory induced by \(\pi\). In the selection rule below, \(\widetilde{\boldsymbol{\delta}}_t(\tau)\) denotes the ridge prediction of surrogate progress for candidate task \(\tau\), and \(\widehat{\boldsymbol{g}}_t^\top\widetilde{\boldsymbol{\delta}}_t(\tau)\) is the corresponding predicted scalar reward.

\section{Method}
\label{sec:method}

SPG-Select begins with \(N\in\mathbb{N}_+\) warmup observations indexed by \(i=1,\ldots,N\), then resets the round index to \(t=1\) for the bandit phase. Warmup samples tasks uniformly within balanced randomized passes. A state-space MIRT model estimates the latent trajectory and item parameters. We train a ridge regression to predict MIRT discrimination vector and difficulty from task embeddings for every task in the task pool used for evolution, including tasks not sampled during warmup. Differences between consecutive approximate profile means smoothed with all warmup data are used to train a multilayer perceptron (MLP) that maps task embeddings and profiles to features for predicting surrogate progress. Online profile estimates instead use predictive means conditioned only on pre-decision observations. During the bandit phase, SPG-Select-R computes the gap weights \(\widehat{\boldsymbol{g}}_t\) from \(\widehat{\boldsymbol{s}}_t\) and scores every task using the fixed MLP encoder and ridge predictions with confidence widths. Each outcome updates the profile and surrogate-progress model, while the trajectory updates reusable backend knowledge. Ridge statistics accumulate within an epoch and reset at the next boundary. In the reported comparisons of Table~\ref{tab:main-results}, SPG-Select-R has higher mean held-out terminal success than SPG-Select-W in three of six backend-benchmark settings, ties in one, and trails in two. We detail SPG-Select-R here and analyze it theoretically.
\begin{figure*}[t]
    \centering
    \includegraphics[width=\textwidth]{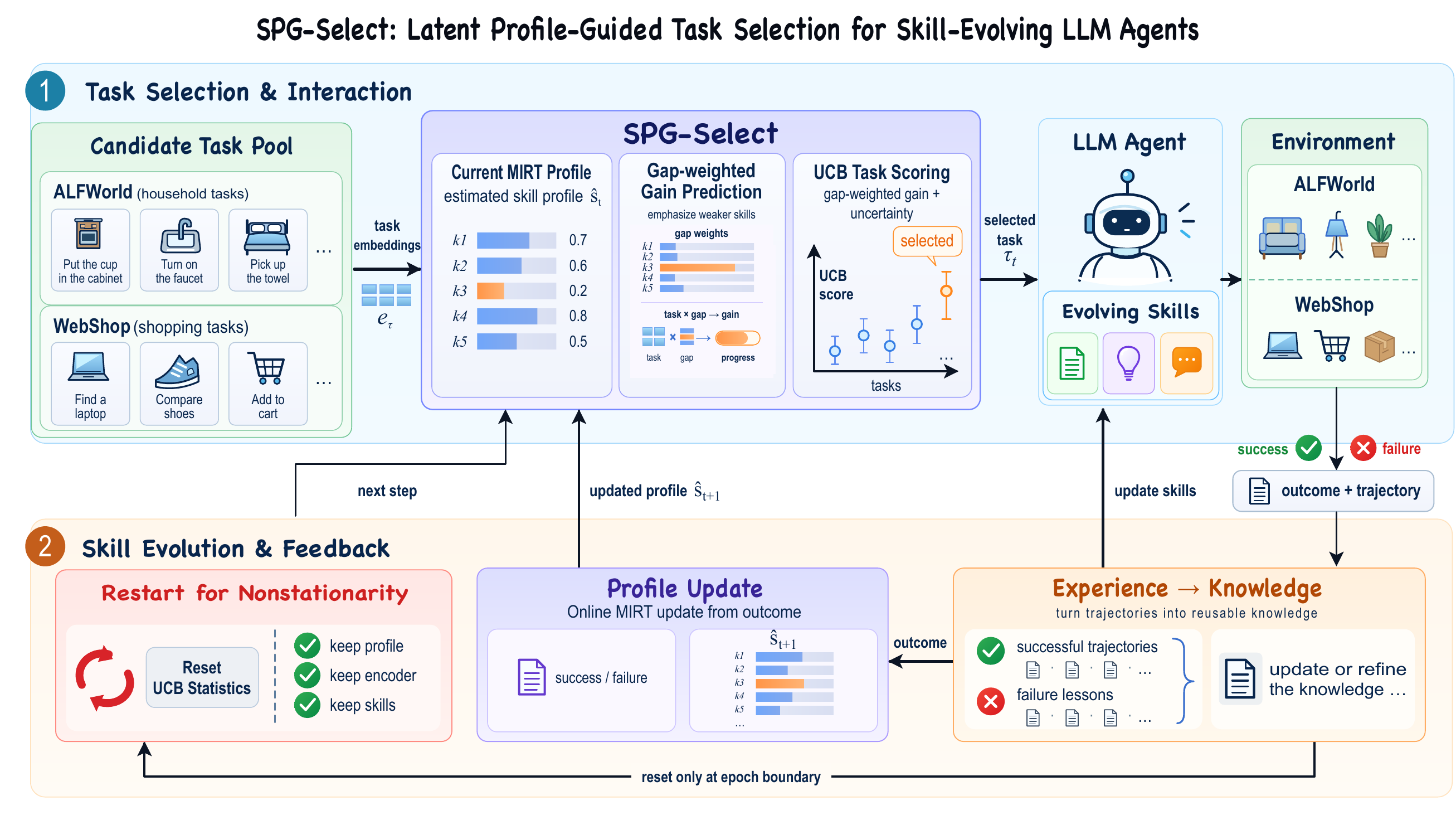}
    \caption{SPG-Select framework for latent profile-guided task selection and skill evolution.}
    \label{fig:framework}
\end{figure*}

\noindent\textbf{MIRT state-space model.}
Each interaction yields a binary outcome, although a task may involve several skills. MIRT uses variation in task loadings across tasks to infer a common \(K\)-dimensional latent state from outcomes \citep{Reckase2006}. We call the predictive mean of this state \(\widehat{\boldsymbol{s}}_t\) the profile, whose coordinates are illustrated schematically in Figure~\ref{fig:framework}. The profile is therefore a model-dependent estimate of latent proficiency rather than an observed decomposition of the agent's skills. MIRT associates task \(\tau\) with an unobserved discrimination vector \(\boldsymbol{a}_\tau^\star\in\mathbb{R}^K\) and difficulty \(d_\tau^\star\in\mathbb{R}\). Given the true latent state \(\boldsymbol{s}_t^\star\), the modeled conditional success probability for task $\tau$ is
\[
    p_\tau^\star(\boldsymbol{s}_t^\star)=\operatorname{sigmoid}\!\left(\boldsymbol{a}_\tau^{\star\top}\boldsymbol{s}_t^\star-d_\tau^\star\right),
\]
where \(\operatorname{sigmoid}(x)=1/(1+e^{-x})\). This Bernoulli likelihood separates task difficulty from state-dependent proficiency. For a generic state \(\boldsymbol{s}\in[0,1]^K\) and item parameters \((\boldsymbol{a}_\tau,d_\tau)\in\mathbb{R}^K\times\mathbb{R}\), let \(p_\tau(\boldsymbol{s})=\operatorname{sigmoid}(\boldsymbol{a}_\tau^\top\boldsymbol{s}-d_\tau)\), and denote the fitted item parameters by \((\widehat{\boldsymbol{a}}_\tau,\widehat d_\tau)\). During warmup, we fit the item parameters and latent trajectory with expectation-maximization (EM) \citep{Dempster1977}. The E-step estimates the latent-state posterior under the current item parameters. The idealized M-step maximizes a regularized expected complete-data log density over the constrained item parameters. For latent states \(\boldsymbol{s}_i\), we use
\(
    p(\boldsymbol{s}_{i+1}\mid\boldsymbol{s}_i)=\mathcal{N}_{[0,1]^K}(\boldsymbol{s}_i,\boldsymbol{\Sigma}_s),
\)
where \(\boldsymbol{\Sigma}_s\in\mathbb{R}^{K\times K}\) is fixed and positive definite, and \(\mathcal{N}_{[0,1]^K}\) denotes a Gaussian distribution truncated to \([0,1]^K\). After warmup, item parameters are fixed and the MIRT belief is filtered online. Let \(b_t(\boldsymbol{s})\) be the predictive density before round \(t\), and let \(\boldsymbol{s}'\in[0,1]^K\) be the next-state variable. Since the likelihood of \(y_t\) is conditioned on \(\boldsymbol{s}\), the update first conditions on \(y_t\) and then propagates the state,
\[
b_t^+(\boldsymbol{s})
\propto p(y_t\mid\boldsymbol{s},\widehat{\boldsymbol{a}}_{\tau_t},\widehat d_{\tau_t})b_t(\boldsymbol{s}),
\qquad
b_{t+1}(\boldsymbol{s}')
=\int_{[0,1]^K}p(\boldsymbol{s}'\mid\boldsymbol{s})b_t^+(\boldsymbol{s})\,d\boldsymbol{s},
\qquad
\widehat{\boldsymbol{s}}_{t+1}=\mathbb{E}_{b_{t+1}}[\boldsymbol{s}'].
\]
Here, \(b_t^+\) is the posterior for the current latent state after observing \(y_t\), and \(b_{t+1}\) predicts the next latent state. Its mean \(\widehat{\boldsymbol{s}}_{t+1}\) is the next profile estimate, and consecutive profiles define surrogate progress. The vector-valued profile makes the dimension-wise gap vector in \eqref{eq:gap-weights} well defined. Appendix~\ref{app:mirt-estimation} details the EM objective, supplementary model properties, and the identification and numerical choices needed to specify the estimator. Algorithm~\ref{alg:bayes-update} in Appendix~\ref{app:algorithms} gives the complete procedure. Notably, the profile interpretation is conditional on the fitted state-space model. 

\noindent\textbf{Feature construction.}
To predict surrogate progress, defined as the temporal difference between consecutive profile estimates, we train a two-layer MLP \(\widehat\Phi\), where \(\operatorname{cat}\) denotes vector concatenation. The learned encoder maps \(\operatorname{cat}(\boldsymbol{e}_\tau,\widehat{\boldsymbol{s}})\) to a unit-ball feature \(\boldsymbol{\phi}\in\mathbb{R}^{d_f}\) via a hidden vector \(\boldsymbol{h}\in\mathbb{R}^{d_h}\), where \(d_h,d_f\in\mathbb{N}_+\). The fitted encoder parameters are \(\widehat{\mathbf{U}}_1\in\mathbb{R}^{d_h\times(d_e+K)}\), \(\widehat{\boldsymbol{c}}_1\in\mathbb{R}^{d_h}\), \(\widehat{\mathbf{U}}_2\in\mathbb{R}^{d_f\times d_h}\), and \(\widehat{\boldsymbol{c}}_2\in\mathbb{R}^{d_f}\). The maximum in the hidden-layer definition is applied coordinatewise:
\begin{align*}
    \boldsymbol{h} = \max\!\left(\widehat{\mathbf{U}}_1\operatorname{cat}(\boldsymbol{e}_\tau,\widehat{\boldsymbol{s}})+\widehat{\boldsymbol{c}}_1,0\right), \qquad
    \widetilde{\boldsymbol{\phi}} &= \widehat{\mathbf{U}}_2 \boldsymbol{h} + \widehat{\boldsymbol{c}}_2,
    \qquad
    \boldsymbol{\phi} = \frac{\widetilde{\boldsymbol{\phi}}}{\max\{1,\lVert\widetilde{\boldsymbol{\phi}}\rVert_2\}}.
\end{align*}
For each warmup index \(i\), let \(\widehat{\boldsymbol{s}}_i\) denote the approximate smoothed profile mean defined in Appendix~\ref{app:mirt-estimation}, and define the pretraining label as \(\widehat{\boldsymbol{\delta}}_i=\widehat{\boldsymbol{s}}_{i+1}-\widehat{\boldsymbol{s}}_i\). These labels are constructed solely from warmup data, whereas online feedback is computed from temporal differences between consecutive predictive means. A linear pretraining head \(\widehat{\mathbf{U}}_3\in\mathbb{R}^{d_f\times K}\) maps \(\boldsymbol{\phi}\) to \(\widehat{\mathbf{U}}_3^\top\boldsymbol{\phi}\) and is trained to predict these labels. The normalization is part of \(\widehat\Phi\), so \(\lVert\widehat\Phi(\cdot)\rVert_2\leq1\) by construction. The encoder is trained on \(\{(\operatorname{cat}(\boldsymbol{e}_{\tau_i},\widehat{\boldsymbol{s}}_i),\widehat{\boldsymbol{\delta}}_i)\}_{i=1}^{N}\). After pretraining, \(\widehat{\mathbf{U}}_3\) is discarded and \(\widehat\Phi\) is fixed.

\noindent\textbf{Gap-weighted UCB selection.}
Periodic restarting is a standard strategy for handling nonstationarity in bandit problems, since it limits the influence of observations generated under earlier parameter regimes \citep{Zhao2020}. Our SPG-Select-R policy extends periodically restarted UCB selection to vector-valued surrogate progress and resets its neural linear ridge statistics at each epoch boundary. Within each epoch, gap-weighted UCB directs practice toward tasks with high predicted surrogate progress in dimensions with larger current profile gaps, while exploring tasks whose contributions remain uncertain. Fix a restart period \(H\in\{1,\ldots,T\}\). Let \(t_0=1+H\lfloor(t-1)/H\rfloor\) be the first round of the epoch containing \(t\), and let \(\mathcal{E}_t=\{t_0,\ldots,t-1\}\). For each \(i\in\mathcal{E}_t\), define \(\boldsymbol{\phi}_i=\widehat\Phi(\operatorname{cat}(\boldsymbol{e}_{\tau_i},\widehat{\boldsymbol{s}}_i))\). With ridge regularization parameter \(\lambda>0\) and \(\mathbf{I}_{d_f}\) denoting the \(d_f\times d_f\) identity matrix, the ridge estimate \(\widehat{\mathbf{W}}_t\in\mathbb{R}^{d_f\times K}\), design matrix \(\mathbf{A}_t\in\mathbb{R}^{d_f\times d_f}\), and response matrix \(\mathbf{B}_t\in\mathbb{R}^{d_f\times K}\) are
\[
    \widehat{\mathbf{W}}_t = \mathbf{A}_t^{-1}\mathbf{B}_t,
    \quad
    \mathbf{A}_t = \lambda \mathbf{I}_{d_f} + \sum_{i \in \mathcal{E}_t} \boldsymbol{\phi}_{i}\boldsymbol{\phi}_{i}^\top,
    \quad
\mathbf{B}_t = \sum_{i \in \mathcal{E}_t} \boldsymbol{\phi}_{i} \widehat{\boldsymbol{\delta}}_{i}^\top.
\]
Since \(\lambda>0\), \(\mathbf{A}_t\succeq\lambda\mathbf{I}_{d_f}\succ0\), where \(\succeq\) denotes the positive-semidefinite order and \(\succ0\) denotes positive definiteness. Thus, $\mathbf{A}_t^{-1}$ is well defined. For a candidate task \(\tau\), let \(\boldsymbol{\phi}_{t,\tau}=\widehat\Phi(\operatorname{cat}(\boldsymbol{e}_\tau,\widehat{\boldsymbol{s}}_t))\). Unlike conventional linear bandits that model directly observed scalar rewards \citep{AbbasiYadkori2011}, our binary observation is converted through the MIRT update into an estimated \(K\)-dimensional profile temporal difference. A task may affect several dimensions whose decision value depends on the current profile. SPG-Select therefore first predicts the vector-valued surrogate progress \(\widetilde{\boldsymbol{\delta}}_t(\tau)=\widehat{\mathbf{W}}_t^\top\boldsymbol{\phi}_{t,\tau}\) and then evaluates it using the current gap vector \(\widehat{\boldsymbol{g}}_t\). Fitting a scalar reward would combine the dimensions using the weights present when each label was formed. Vector prediction instead retains the predicted temporal difference in every profile dimension, allowing the same fitted surrogate-progress model to be evaluated using the current weights as the profile evolves. Write \(w_t(\tau)=\lVert\boldsymbol{\phi}_{t,\tau}\rVert_{\mathbf{A}_t^{-1}}=\sqrt{\boldsymbol{\phi}_{t,\tau}^\top\mathbf{A}_t^{-1}\boldsymbol{\phi}_{t,\tau}}\) for the confidence width. The UCB score is
\[
    \mathcal{U}_t(\tau) = \widehat{\boldsymbol{g}}_t^\top \widetilde{\boldsymbol{\delta}}_t(\tau) + \beta_t\lVert\widehat{\boldsymbol{g}}_t\rVert_2w_t(\tau).
\]
The policy selects a maximizer of \(\mathcal{U}_t(\tau)\), with ties broken deterministically. Its first term, labeled ``gain'' in Figure~\ref{fig:framework}, is the predicted gap-weighted surrogate progress \(\widehat{\boldsymbol{g}}_t^\top\widetilde{\boldsymbol{\delta}}_t(\tau)\). Its second term is a confidence bonus for this scalarized prediction, where \(\beta_t>0\) controls exploration. The theoretical guarantee uses the calibrated radius in \eqref{eq:theory-beta}, whereas the reported experiments use the fixed coefficient specified in Table~\ref{tab:selector-hyperparameters}. Given the fitted components and prescribed control parameters, the predictive density \(b_t\) and ridge matrices \(\mathbf{A}_t,\mathbf{B}_t\) constitute the state used by the selector at round \(t\). They determine the current profile estimate, surrogate-progress predictions, and confidence widths, and are updated recursively after each observation, with the ridge statistics reset at epoch boundaries.

\noindent\textbf{Algorithm overview.}
Algorithms~\ref{alg:spg} and~\ref{alg:bayes-update} in Appendix~\ref{app:algorithms} give the complete procedures for SPG-Select-R and the profile update. At a high level, warmup calibrates the MIRT estimator and trains the task-profile representation that is fixed thereafter. Each online round converts the pre-decision profile estimate into gap weights, selects the task with the largest surrogate-progress UCB score, and uses the observed outcome to compute the next profile estimate and its temporal difference for the current epoch's ridge update. At an epoch boundary, only the ridge/UCB statistics are reset. The MIRT estimates, frozen encoder, and backend state are retained. Thus, restarting limits the influence of stale surrogate-progress observations while preserving the accumulated skill-evolution state.

\section{Experiments}
\label{sec:experiments}

Our experiments examine whether profile-guided task selection improves held-out performance across skill-evolution mechanisms, how SPG-Select-R compares with SPG-Select-W, and how UCB exploration and profile-gap weighting affect performance. We study ALFWorld and WebShop with SkillRL, ExpeL, and Shared Reflexion.

\subsection{Experimental Setup}

\noindent\textbf{Benchmarks.}
ALFWorld \citep{Shridhar2020} is a text-based household environment in which agents complete tasks involving object manipulation and compositional action sequences. We evaluate on 134 tasks from its unseen split, comprising 24 Pick, 18 Look, 31 Clean, 23 Heat, 21 Cool, and 17 Pick2 tasks. This evaluation tests whether acquired knowledge transfers to unseen household situations. WebShop \citep{Yao2022} is a simulated shopping environment that requires agents to search for products, inspect their attributes, and purchase an item satisfying a natural-language instruction. We use its 1,000-task development pool for skill evolution and its 500-task test pool for evaluation. Together, the benchmarks cover household planning and goal-directed web interaction. Additional implementation details, computational resources, and backend-specific hyperparameters are reported in Appendix~\ref{app:experimental-details}.

\noindent\textbf{Practice and evaluation.}
During practice, the task-selection policy selects a task and the agent makes one attempt using its current backend state. For each backend-benchmark comparison, every policy first receives the same \(N\) warmup interactions. The backend state produced by this common warmup initializes the ensuing \(T\)-round policy comparison. SPG-Select uses the observed outcome to compute the surrogate-progress feedback defined in Section~\ref{sec:method}, namely the temporal difference between consecutive profile estimates. This feedback is computed from the outcome before the trajectory is used for the subsequent backend update. Actor-model parameters remain fixed, so adaptation takes place through accumulated skills, insights, or reflections. After practice, we freeze the resulting backend state and evaluate each held-out task with a single attempt. No updates or test-time reflection retries are performed during evaluation. This protocol measures the transfer of acquired knowledge to held-out tasks.

\noindent\textbf{Metrics.}
We report the terminal task success rate as a percentage of the evaluation pool. ALFWorld success is determined by the environment's task-completion signal. WebShop success requires a terminal reward of \(1.0\), and partial rewards do not count as successes. For WebShop, we also report Score, the mean terminal reward expressed as a percentage, which preserves partial credit. We additionally report ALFWorld results by task family to examine variation across task types. All reported gains are percentage-point differences computed from unrounded rates.

\noindent\textbf{Random seeds.}
The main results in Table~\ref{tab:main-results} are reported as the mean and sample standard deviation across three random seeds. All other experimental figures and tables in the main text and appendix use a single fixed random seed shared across the compared methods.

\noindent\textbf{Skill-evolution backends.}
We integrate the skill-evolution component of three backends. SkillRL contributes its hierarchical SkillBank and failure-driven skill acquisition \citep{Xia2026}. ExpeL contributes experience-derived insights and demonstration retrieval \citep{Zhao2024}. Shared Reflexion is our cross-task memory adaptation of the original Reflexion method \citep{Shinn2023}. It replaces task-local episodic memory with a cross-task memory. Representative persisted knowledge and held-out execution traces appear in Appendix~\ref{app:qualitative-examples}.

\noindent\textbf{Task-selection policies.}
In the tables, Uniform denotes uniform task selection without a bandit model. It samples without replacement from the full task pool within each pass and draws a new random ordering for the next pass, independently of observed outcomes or backend state. The sliding-window variant SPG-Select-W is described in Appendix~\ref{app:sliding-window}. Comparing SPG-Select-R with SPG-Select-W evaluates two approaches to nonstationary adaptive task selection.

\noindent\textbf{Implementation fidelity.}
Within each backend-benchmark comparison, the task-selection policies share the practice and held-out task pools, backend update mechanism, initial skill pool, and random seed. The integrated implementation also supports backend-state features, task-eligibility rules, and backend-specific stopping conditions, as described in Appendix~\ref{app:backend-integration}.

\subsection{Results}

Table~\ref{tab:main-results} compares the task-selection policies across backends and benchmarks, with ALFWorld results reported by task family. We report the mean and sample standard deviation across three random seeds, using the same three seeds for every backend, benchmark, and policy. Each seed is evaluated on all 134 ALFWorld tasks or all 500 WebShop tasks. Standard deviations describe variation across seeds and are not standard errors.
In both tables, `All' denotes the task-count-weighted ALFWorld success rate. For ALFWorld All and the two WebShop metrics in Table~\ref{tab:main-results}, parenthetical values give percentage-point changes in the mean from Uniform, computed from unrounded values for the same backend and metric. Bold and underlined entries mark the best and second-best means within each backend and metric, with ties receiving the same formatting. Table~\ref{tab:ablations} separately reports the single-seed ablations.

\begin{table*}[t]
\centering
\footnotesize
\setlength{\tabcolsep}{2pt}
\renewcommand{\arraystretch}{1.28}
\definecolor{mainrow}{RGB}{232,242,255}
\caption{Held-out performance (\%) of task-selection policies on ALFWorld and WebShop with three skill-evolution backends. Values are mean $\pm$ sample standard deviation across three random seeds. ALFWorld All is task-count-weighted overall success. WebShop Score is mean terminal reward and Succ. is success rate. Light blue shading highlights SPG-Select-W and SPG-Select-R. Within each backend and metric, \textbf{bold} and \underline{underlined} means indicate the best and second-best results. Ties share the same formatting. Parenthetical values in All, Score, and Succ. give percentage-point changes from Uniform. Higher is better for all metrics.}
\label{tab:main-results}
\resizebox{\linewidth}{!}{%
\begin{tabular}{@{}llccccccc@{\hspace{5pt}}|@{\hspace{5pt}}cc@{}}
\hline
\multirow{2}{*}{Backend} & \multirow{2}{*}{Selection policy} & \multicolumn{7}{c@{\hspace{5pt}}|@{\hspace{5pt}}}{ALFWorld} & \multicolumn{2}{c}{WebShop} \\
\cline{3-11}
& & Pick & Look & Clean & Heat & Cool & Pick2 & All & Score & Succ. \\
\hline
\multirow{3}{*}{SkillRL}
& Uniform & 47.22{\scriptsize$\pm\,8.67$} & \underline{74.07}{\scriptsize$\pm\,8.49$} & \textbf{55.91}{\scriptsize$\pm\,8.12$} & \textbf{71.01}{\scriptsize$\pm\,5.02$} & \textbf{58.73}{\scriptsize$\pm\,2.75$} & \underline{39.22}{\scriptsize$\pm\,3.40$} & \underline{57.71}{\scriptsize$\pm\,4.97$} & 61.76{\scriptsize$\pm\,2.04$} & 34.67{\scriptsize$\pm\,1.33$} \\
& \cellcolor{mainrow}\textbf{SPG-Select-W} & \cellcolor{mainrow}\underline{50.00}{\scriptsize$\pm\,8.33$} & \cellcolor{mainrow}\textbf{77.78}{\scriptsize$\pm\,0.00$} & \cellcolor{mainrow}52.69{\scriptsize$\pm\,15.24$} & \cellcolor{mainrow}\underline{66.67}{\scriptsize$\pm\,2.51$} & \cellcolor{mainrow}\underline{53.97}{\scriptsize$\pm\,9.91$} & \cellcolor{mainrow}\textbf{43.14}{\scriptsize$\pm\,8.99$} & \cellcolor{mainrow}56.97{\scriptsize$\pm\,5.50$}{\tiny\,(-0.75)} & \cellcolor{mainrow}\underline{63.47}{\scriptsize$\pm\,0.88$}{\tiny\,(+1.71)} & \cellcolor{mainrow}\textbf{37.33}{\scriptsize$\pm\,0.81$}{\tiny\,(+2.67)} \\
& \cellcolor{mainrow}\textbf{SPG-Select-R} & \cellcolor{mainrow}\textbf{59.72}{\scriptsize$\pm\,6.36$} & \cellcolor{mainrow}\underline{74.07}{\scriptsize$\pm\,11.56$} & \cellcolor{mainrow}\underline{54.84}{\scriptsize$\pm\,8.53$} & \cellcolor{mainrow}\textbf{71.01}{\scriptsize$\pm\,2.51$} & \cellcolor{mainrow}52.38{\scriptsize$\pm\,4.76$} & \cellcolor{mainrow}\textbf{43.14}{\scriptsize$\pm\,6.79$} & \cellcolor{mainrow}\textbf{59.20}{\scriptsize$\pm\,0.43$}{\tiny\,(+1.49)} & \cellcolor{mainrow}\textbf{64.19}{\scriptsize$\pm\,1.06$}{\tiny\,(+2.43)} & \cellcolor{mainrow}\underline{36.60}{\scriptsize$\pm\,1.06$}{\tiny\,(+1.93)} \\
\hline
\multirow{3}{*}{ExpeL}
& Uniform & 19.44{\scriptsize$\pm\,9.62$} & 29.63{\scriptsize$\pm\,16.97$} & \underline{35.48}{\scriptsize$\pm\,8.53$} & 27.54{\scriptsize$\pm\,17.57$} & 36.51{\scriptsize$\pm\,11.98$} & 0.00{\scriptsize$\pm\,0.00$} & 26.12{\scriptsize$\pm\,3.73$} & \underline{56.00}{\scriptsize$\pm\,3.66$} & 19.27{\scriptsize$\pm\,1.80$} \\
& \cellcolor{mainrow}\textbf{SPG-Select-W} & \cellcolor{mainrow}\textbf{36.11}{\scriptsize$\pm\,4.81$} & \cellcolor{mainrow}\underline{59.26}{\scriptsize$\pm\,11.56$} & \cellcolor{mainrow}34.41{\scriptsize$\pm\,1.86$} & \cellcolor{mainrow}\textbf{37.68}{\scriptsize$\pm\,9.05$} & \cellcolor{mainrow}\textbf{58.73}{\scriptsize$\pm\,2.75$} & \cellcolor{mainrow}\textbf{3.92}{\scriptsize$\pm\,3.40$} & \cellcolor{mainrow}\underline{38.56}{\scriptsize$\pm\,3.02$}{\tiny\,(+12.44)} & \cellcolor{mainrow}\textbf{56.03}{\scriptsize$\pm\,0.65$}{\tiny\,(+0.03)} & \cellcolor{mainrow}\textbf{20.60}{\scriptsize$\pm\,3.17$}{\tiny\,(+1.33)} \\
& \cellcolor{mainrow}\textbf{SPG-Select-R} & \cellcolor{mainrow}\underline{34.72}{\scriptsize$\pm\,2.41$} & \cellcolor{mainrow}\textbf{62.96}{\scriptsize$\pm\,11.56$} & \cellcolor{mainrow}\textbf{40.86}{\scriptsize$\pm\,4.93$} & \cellcolor{mainrow}\underline{34.78}{\scriptsize$\pm\,4.35$} & \cellcolor{mainrow}\underline{57.14}{\scriptsize$\pm\,9.52$} & \cellcolor{mainrow}\underline{1.96}{\scriptsize$\pm\,3.40$} & \cellcolor{mainrow}\textbf{39.30}{\scriptsize$\pm\,4.97$}{\tiny\,(+13.18)} & \cellcolor{mainrow}47.18{\scriptsize$\pm\,4.49$}{\tiny\,(-8.82)} & \cellcolor{mainrow}\underline{20.20}{\scriptsize$\pm\,2.23$}{\tiny\,(+0.93)} \\
\hline
\multirow{3}{*}{\shortstack[l]{Shared\\Reflexion}}
& Uniform & 25.00{\scriptsize$\pm\,7.22$} & \underline{53.70}{\scriptsize$\pm\,8.49$} & 30.11{\scriptsize$\pm\,9.86$} & \underline{30.43}{\scriptsize$\pm\,11.50$} & 36.51{\scriptsize$\pm\,5.50$} & 3.92{\scriptsize$\pm\,6.79$} & 30.10{\scriptsize$\pm\,4.97$} & 39.72{\scriptsize$\pm\,13.88$} & \underline{18.27}{\scriptsize$\pm\,6.20$} \\
& \cellcolor{mainrow}\textbf{SPG-Select-W} & \cellcolor{mainrow}\textbf{31.94}{\scriptsize$\pm\,8.67$} & \cellcolor{mainrow}48.15{\scriptsize$\pm\,3.21$} & \cellcolor{mainrow}\underline{33.33}{\scriptsize$\pm\,7.45$} & \cellcolor{mainrow}\textbf{40.58}{\scriptsize$\pm\,9.05$} & \cellcolor{mainrow}\textbf{50.79}{\scriptsize$\pm\,16.72$} & \cellcolor{mainrow}\underline{5.88}{\scriptsize$\pm\,10.19$} & \cellcolor{mainrow}\underline{35.57}{\scriptsize$\pm\,7.98$}{\tiny\,(+5.47)} & \cellcolor{mainrow}\textbf{48.86}{\scriptsize$\pm\,3.49$}{\tiny\,(+9.15)} & \cellcolor{mainrow}\textbf{21.40}{\scriptsize$\pm\,2.51$}{\tiny\,(+3.13)} \\
& \cellcolor{mainrow}\textbf{SPG-Select-R} & \cellcolor{mainrow}\underline{30.56}{\scriptsize$\pm\,4.81$} & \cellcolor{mainrow}\textbf{57.41}{\scriptsize$\pm\,6.42$} & \cellcolor{mainrow}\textbf{43.01}{\scriptsize$\pm\,7.45$} & \cellcolor{mainrow}24.64{\scriptsize$\pm\,15.27$} & \cellcolor{mainrow}\underline{46.03}{\scriptsize$\pm\,2.75$} & \cellcolor{mainrow}\textbf{13.73}{\scriptsize$\pm\,3.40$} & \cellcolor{mainrow}\textbf{36.32}{\scriptsize$\pm\,3.02$}{\tiny\,(+6.22)} & \cellcolor{mainrow}\underline{48.31}{\scriptsize$\pm\,2.14$}{\tiny\,(+8.59)} & \cellcolor{mainrow}\textbf{21.40}{\scriptsize$\pm\,0.53$}{\tiny\,(+3.13)} \\
\hline
\end{tabular}
}
\end{table*}
\begin{figure*}[t]
    \centering
    \begin{subfigure}[t]{0.49\textwidth}
        \centering
        \includegraphics[width=\linewidth]{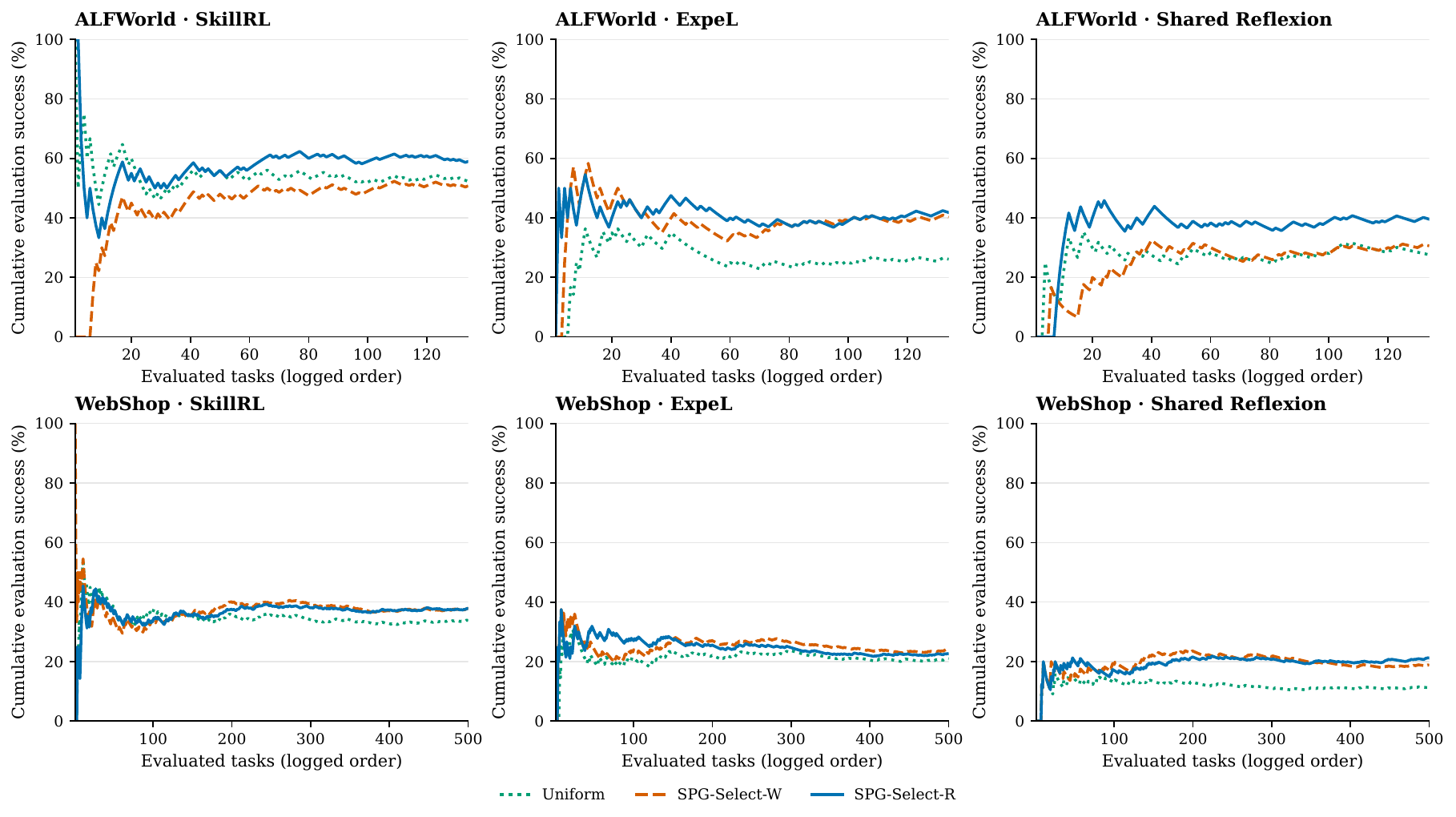}
        \caption{Cumulative held-out success over a common evaluation order.}
        \label{fig:cumulative-evaluation-success}
    \end{subfigure}\hfill
    \begin{subfigure}[t]{0.49\textwidth}
        \centering
        \includegraphics[width=\linewidth]{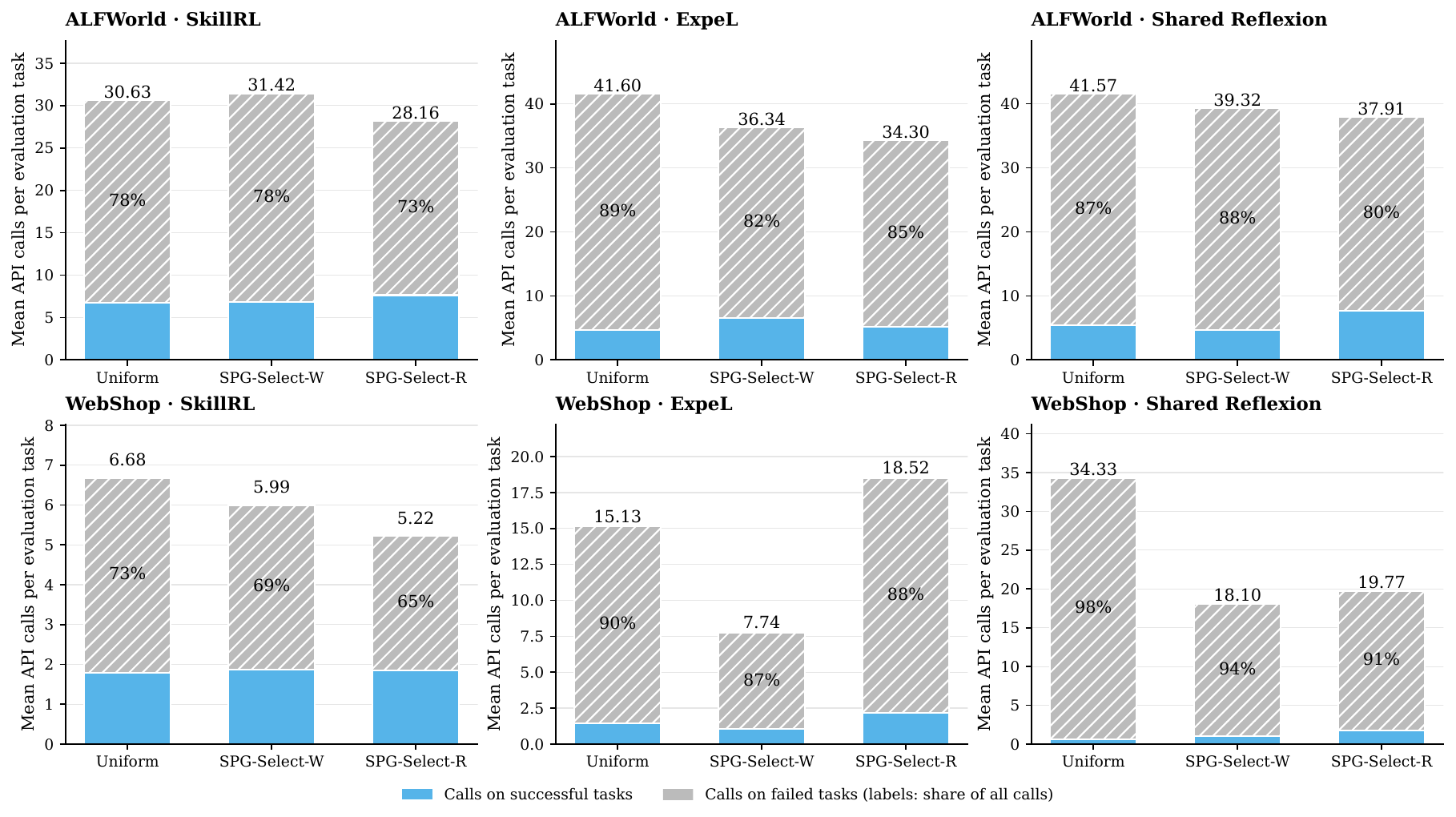}
        \caption{Mean application programming interface (API) calls per held-out task, decomposed by outcome.}
        \label{fig:mean-evaluation-api-calls}
    \end{subfigure}
    \caption{Held-out evaluation behavior with frozen backend memories.}
    \label{fig:evaluation-diagnostics}
\end{figure*}

Across the three reported seeds, SPG-Select-R exceeds Uniform in mean success in all six backend-benchmark combinations, with gains of 1.49 to 13.18 percentage points on ALFWorld and 0.93 to 3.13 percentage points on WebShop. SPG-Select-W exceeds Uniform in mean success in five of the six settings; for SkillRL on ALFWorld, its mean is 0.75 percentage points lower. SPG-Select-R achieves higher mean success than SPG-Select-W for all three backends on ALFWorld. On WebShop, SPG-Select-W exceeds SPG-Select-R by 0.73 points for SkillRL and 0.40 points for ExpeL, while both policies achieve 21.40\% mean success for Shared Reflexion. Thus, SPG-Select-R has higher mean success than SPG-Select-W in three settings, ties in one, and trails in two. WebShop Score follows a different pattern: SPG-Select-R has the highest mean Score for SkillRL, whereas SPG-Select-W has the highest mean Score for ExpeL and Shared Reflexion. For ExpeL, the mean Scores of SPG-Select-W and Uniform are nearly equal (56.03\% and 56.00\%), while SPG-Select-R has a lower mean Score despite higher mean terminal success than Uniform. Task-family rankings vary within ALFWorld. Relative to SPG-Select-W, SPG-Select-R has its largest mean gain on Pick for SkillRL and on Clean for Shared Reflexion, whereas SPG-Select-W achieves higher mean success on Heat for ExpeL and Shared Reflexion.

Figure~\ref{fig:cumulative-evaluation-success} reports cumulative success over a common logged evaluation order for the single-seed reference runs. Since backend states are frozen, changes across prefixes reflect the task prefixes rather than test-time updates. Figure~\ref{fig:mean-evaluation-api-calls} decomposes mean API calls into contributions from success and failures, using the full evaluation-pool size as the denominator for both components. Failed tasks account for most calls, especially for ExpeL and Shared Reflexion. These costs describe the frozen agents and do not measure practice-time sample efficiency. Appendix~\ref{app:api-calls} reports total API calls and comparisons with Uniform.
Appendix~\ref{app:surrogate-practice-association} shows two sets of experiments that together support the use of gap-weighted reward as a feedback signal for skill evolution.

\subsection{Ablation Study}

Table~\ref{tab:ablations} combines ALFWorld component ablations with profile-dimension and history ablations using a single fixed random seed. All ablation variants and their baselines use the same fixed random seed. Every ablation uses exactly the same practice and held-out task pools as its corresponding main experiment, so the comparisons are not affected by differences in task-pool composition. Here, w/o UCB means without the UCB bonus, and w/o gap means without gap weighting. w/o UCB sets \(\beta_t=0\), removing the confidence bonus while retaining the gap-weighted prediction. w/o gap sets \(\widehat{\boldsymbol{g}}_t=\mathbf{1}/K\) in the selection score while retaining \(\beta_t=0.1\). It therefore weights skill dimensions equally while preserving the confidence bonus and the estimated profile in the context. Relative to SPG-Select-R, these ablations isolate the confidence bonus and gap weighting, respectively. The direct comparison between w/o UCB and w/o gap differs in both components. The \(K=1\) setting uses a scalar latent profile with periodic restarting, whereas Full history uses the \(K=4\) profile while retaining all ridge observations. Parentheses in the All column indicate percentage-point changes from SPG-Select-R, computed from unrounded rates.
\begin{table*}[t]
\centering
\fontsize{8.5}{10.2}\selectfont
\setlength{\tabcolsep}{4pt}
\renewcommand{\arraystretch}{1.08}
\caption{ALFWorld ablations (\%). All deltas vs. SPG-Select-R.}
\label{tab:ablations}
\begin{tabular*}{0.92\linewidth}{@{\extracolsep{\fill}}llccccccc@{}}
\toprule
Backend & Setting & Pick & Look & Clean & Heat & Cool & Pick2 & All \\
\midrule
\multirow{6}{*}{SkillRL}
& Uniform & 37.5 & 66.7 & \underline{48.4} & 65.2 & \underline{57.1} & \underline{41.2} & 52.2{\tiny\,(-6.7)} \\
& w/o UCB & 50.0 & 66.7 & 38.7 & 65.2 & \textbf{61.9} & 35.3 & 52.2{\tiny\,(-6.7)} \\
& w/o gap & 50.0 & \textbf{77.8} & \textbf{54.8} & 65.2 & \underline{57.1} & 35.3 & \underline{56.7}{\tiny\,(-2.2)} \\
& $K=1$ & 50.0 & \underline{72.2} & \textbf{54.8} & \textbf{73.9} & 52.4 & 35.3 & \underline{56.7}{\tiny\,(-2.2)} \\
& Full history & \underline{58.3} & 61.1 & 41.9 & \underline{69.6} & 47.6 & \underline{41.2} & 53.0{\tiny\,(-6.0)} \\
& \textbf{SPG-Select-R} & \textbf{66.7} & \textbf{77.8} & 45.2 & \underline{69.6} & 52.4 & \textbf{47.1} & \textbf{59.0} \\
\midrule
\multirow{6}{*}{ExpeL}
& Uniform & 8.3 & 44.4 & \underline{41.9} & 30.4 & 23.8 & 0.0 & 26.1{\tiny\,(-15.7)} \\
& w/o UCB & \underline{33.3} & 44.4 & 35.5 & \underline{52.2} & \underline{47.6} & 0.0 & 36.6{\tiny\,(-5.2)} \\
& w/o gap & \underline{33.3} & 38.9 & 35.5 & \textbf{56.5} & \textbf{57.1} & \textbf{17.6} & 40.3{\tiny\,(-1.5)} \\
& $K=1$ & 29.2 & \textbf{83.3} & 29.0 & 43.5 & 19.0 & 0.0 & 33.6{\tiny\,(-8.2)} \\
& Full history & \textbf{37.5} & 55.6 & \textbf{45.2} & 39.1 & \textbf{57.1} & \underline{5.9} & \underline{41.0}{\tiny\,(-0.7)} \\
& \textbf{SPG-Select-R} & \textbf{37.5} & \underline{66.7} & \textbf{45.2} & 39.1 & \textbf{57.1} & 0.0 & \textbf{41.8} \\
\midrule
\multirow{6}{*}{\shortstack[l]{Shared\\Reflexion}}
& Uniform & 16.7 & \underline{55.6} & \underline{38.7} & 17.4 & 33.3 & 0.0 & 27.6{\tiny\,(-11.9)} \\
& w/o UCB & \underline{33.3} & \textbf{61.1} & 29.0 & \textbf{39.1} & \underline{42.9} & \underline{5.9} & \underline{35.1}{\tiny\,(-4.5)} \\
& w/o gap & 8.3 & \underline{55.6} & 19.4 & 17.4 & 28.6 & 0.0 & 20.9{\tiny\,(-18.7)} \\
& $K=1$ & 25.0 & 50.0 & 25.8 & \underline{34.8} & \underline{42.9} & 0.0 & 29.9{\tiny\,(-9.7)} \\
& Full history & \textbf{37.5} & 50.0 & \underline{38.7} & 13.0 & \textbf{47.6} & 0.0 & 32.1{\tiny\,(-7.5)} \\
& \textbf{SPG-Select-R} & 25.0 & \textbf{61.1} & \textbf{51.6} & \textbf{39.1} & \underline{42.9} & \textbf{11.8} & \textbf{39.6} \\
\bottomrule
\end{tabular*}
\end{table*}

On ALFWorld, SPG-Select-R improves aggregate success over w/o UCB by 6.7, 5.2, and 4.5 percentage points for SkillRL, ExpeL, and Shared Reflexion. Its gains over w/o gap are 2.2, 1.5, and 18.7 points, respectively. Relative to the scalar-profile \(K=1\) setting, the gains are 2.2, 8.2, and 9.7 points. Relative to Full history, they are 6.0, 0.7, and 7.5 points. The w/o UCB and w/o gap results are consistent with benefits from uncertainty-driven exploration and profile-gap weighting, respectively. The lower aggregate success under \(K=1\) and Full history is consistent with benefits from a multidimensional profile and periodically resetting the ridge statistics. The task-family results vary, but SPG-Select-R has the highest aggregate success for each backend. We also evaluate the ablations on WebShop, where SPG-Select-R achieves higher success than the ablations. To keep the main text concise, detailed results and metric-level analysis appear in Appendix~\ref{app:experimental-details} (Table~\ref{tab:webshop-ablations}).

\section{Theoretical Analysis}
\label{sec:theory}

We analyze vector-valued SPG-Select-R under path variation, following the analysis of \citet{Zhao2020}. All statements condition on the warmup outputs, with the task pool \(\mathcal{T}\), encoder \(\widehat\Phi\), and MIRT item parameters fixed. The model and noise assumptions in Appendix~\ref{app:proof}, which also provides the proofs, must hold under this conditioning. The guarantee concerns surrogate feedback under exact linear realizability, which is a separate modeling assumption beyond encoder pretraining. Let \(\mathcal{F}_{t-1}\) denote the sigma-field generated by the warmup outputs and all tasks, outcomes, trajectories, backend updates, and estimator randomness available before round \(t\). For each candidate \(\tau\), let \(\widehat{\boldsymbol{\delta}}_t(\tau)\) be its potential surrogate progress, i.e., the potential temporal difference of the profile estimate. Only \(\widehat{\boldsymbol{\delta}}_t(\tau_t)\) is observed. Define \(\boldsymbol{\phi}_{t,\tau}=\widehat\Phi(\operatorname{cat}(\boldsymbol{e}_\tau,\widehat{\boldsymbol{s}}_t))\) and the conditional mean surrogate progress
\(
\boldsymbol{\mu}_t(\tau)=\mathbb{E}[\widehat{\boldsymbol{\delta}}_t(\tau)\mid\mathcal{F}_{t-1}].
\)
The conditional mean scalar reward is \(q_t(\tau)=\widehat{\boldsymbol{g}}_t^\top\boldsymbol{\mu}_t(\tau)\).
\begin{definition}
\label{def:dynamic-pseudoregret}
Let \(\tau_t^*\in\arg\max_{\tau\in\mathcal{T}}q_t(\tau)\) be a dynamic oracle action. The dynamic pseudo-regret of a deterministic policy over \(T\) bandit rounds is defined as
\[
    R_T=\sum_{t=1}^{T}\bigl[q_t(\tau_t^*)-q_t(\tau_t)\bigr].
\]
\end{definition}
Both the oracle action and the selected action are evaluated along the learner's realized history at the same current profile and backend state. Thus, \(R_T\) measures cumulative one-step allocation quality along the realized history. Under Assumptions~\ref{assump:linear-surrogate} and~\ref{assump:boundedness}, the conditional mean model is \(\boldsymbol{\mu}_t(\tau)=\mathbf{W}_t^{\star\top}\boldsymbol{\phi}_{t,\tau}\), with \(\mathbf{W}_t^\star\in\mathbb{R}^{d_f\times K}\). Definition~\ref{def:path-variation} gives its path length as \(P_T=\sum_{t=1}^{T-1}\lVert\mathbf{W}_{t+1}^\star-\mathbf{W}_t^\star\rVert_F\). Restarting restricts the ridge statistics to the current epoch. For fixed \(H\in\{1,\ldots,T\}\) and \(\rho\in(0,1)\), the confidence radius \(\beta_t\) is defined in \eqref{eq:theory-beta}, and \(\beta=\max_{1\leq t\leq T}\beta_t\).
\begin{theorem}
\label{thm:dynamic-pseudoregret}
Under Assumptions~\ref{assump:linear-surrogate} and~\ref{assump:boundedness}, with path length \(P_T\) as in Definition~\ref{def:path-variation}, fix \(H\in\{1,\ldots,T\}\) before the bandit phase, let \(\rho\in(0,1)\), and define \(\beta_t\) and \(\beta\) by~\eqref{eq:theory-beta}. Then, conditional on the warmup outputs, with probability at least \(1-\rho\), the deterministic UCB policy specified in Algorithm~\ref{alg:spg} of Appendix~\ref{app:algorithms} satisfies
\[
R_T \leq 2\beta\left\lceil\frac{T}{H}\right\rceil\sqrt{\frac{(\lambda+1)d_fH}{\lambda}\log\!\left(1+\frac{H}{\lambda d_f}\right)}+2\sqrt{\frac{d_f}{\lambda}}H^{3/2}P_T.
\]
\end{theorem}
Lemma~\ref{lem:epoch-confidence} bounds prediction error by statistical uncertainty and within-epoch drift, simultaneously for all candidate tasks and rounds on one event of probability at least \(1-\rho\). This uniformity permits adaptive task selection. The selection score uses the statistical bonus, while the proof accounts separately for drift. The theorem's first term is the estimation cost accumulated across restarted epochs, and its second is the bias from within-epoch parameter drift. The bound holds for any fixed restart period and uses the realized, potentially random path length \(P_T\). For oracle tuning, Corollary~\ref{cor:restart-selection} instead assumes a deterministic budget \(V_T\geq0\), specified before the bandit phase, such that \(P_T\leq V_T\) almost surely under the resulting policy. Let \(\beta^{\mathrm{u}}\) be the horizon-uniform radius in \eqref{eq:horizon-uniform-radius}. For \(V_T>0\), define
\(
H^\circ=(\beta^{\mathrm{u}}T\sqrt{(\lambda+1)\log(1+T/(\lambda d_f))}/V_T)^{1/2},
H^*=\min\{T,\max\{1,\lceil H^\circ\rceil\}\}.
\)
Run the policy with \(H=H^*\) and the radii in \eqref{eq:theory-beta}. If \(1\leq H^\circ\leq T\), the corollary gives, conditional on the warmup outputs, with probability at least \(1-\rho\),
\(
R_T=\mathcal{O}\!(d_f^{1/2}(\lambda+1)^{3/8}\log^{3/8}(1+T/(\lambda d_f))(\beta^{\mathrm{u}}T)^{3/4}V_T^{1/4}/\lambda^{1/2}).
\)
Let \(S\) and \(\sigma\) denote the parameter-norm bound and noise scale in Assumption~\ref{assump:boundedness}. When \(K,\lambda,\lambda^{-1},S,\sigma,\rho^{-1}\) are fixed independently of \(d_f\) and \(T\), this becomes \(R_T=\widetilde{\mathcal{O}}(d_f^{7/8}T^{3/4}V_T^{1/4})\). For \(V_T=0\), choosing \(H=T\) gives \(R_T=\widetilde{\mathcal{O}}(d_f\sqrt T)\) in the same regime. The oracle choice uses an a priori variation budget, and substituting the future realized \(P_T\) would not define a nonanticipating policy. The detailed discussion and proof appear in Appendix~\ref{app:proof}.

\section{Conclusion}

We present SPG-Select for finite-budget task selection in skill-evolving LLM agents. It combines MIRT profile estimation, gap-weighted surrogate-progress prediction, and uncertainty-aware selection. Across the three reported seeds, SPG-Select-R achieves higher mean held-out success than Uniform in all six backend-benchmark combinations. Under the stated surrogate model, it admits a high-probability dynamic pseudo-regret bound for surrogate task allocation.

\bibliographystyle{plainnat}
\bibliography{ref_google_scholar}

\clearpage
\appendix
\raggedbottom
\newpage
{\centering
{\Large \textbf{Appendix}}
\par}
\vspace{0.6cm}
{\hypersetup{linkcolor=black}
\tableofcontents
}
\addtocontents{toc}{\protect\setcounter{tocdepth}{2}}

\newpage
\section{Related Work}
\label{app:related-work}

\noindent\textbf{Experience-based skill evolution.}
Experience-based LLM agents can improve behavior without necessarily modifying model parameters. ReAct (reasoning and acting) provides an interface that interleaves these processes \citep{Yao2022a}, Reflexion uses linguistic self-feedback \citep{Shinn2023}, ExpeL distills transferable insights \citep{Zhao2024}, and Voyager couples a generated curriculum with an executable skill library \citep{Wang2023}. Recent work further studies experience sharing, refinement, recursive evolution, and learnable memory skills \citep{Tang2025,Wu2025,Xia2026,Zhang2026}. SPG-Select complements these knowledge-acquisition mechanisms with profile-conditioned practice-task selection. Voyager also selects tasks, using an LLM-generated curriculum rather than the surrogate-progress bandit considered here.

\noindent\textbf{MIRT-based ability estimation.}
MIRT represents ability with a latent vector and tasks with dimension-specific discrimination parameters \citep{Reckase2006}. Scalable item response theory (IRT) estimation \citep{Frick2024} and IRT-based language-model evaluation \citep{Truong2025} extend this perspective to modern machine-learning settings. BanditCAT combines IRT with a contextual bandit for computerized adaptive testing, using Fisher information as the item-selection reward \citep{Sharpnack2024}. In contrast, SPG-Select filters a profile, defined as the MIRT latent-state estimate, and uses its temporal difference as surrogate-progress feedback for practice-task selection.

\noindent\textbf{Contextual and nonstationary bandits.}
Contextual bandits use side information and bandit feedback for sequential action choice \citep{AbbasiYadkori2011}. In contrast, prioritized experience replay selects stored transitions \citep{Schaul2015}. For nonstationarity, discounted UCB \citep{Garivier2008}, sliding-window UCB \citep{Cheung2019}, weighted estimation \citep{Russac2019}, and restarted UCB \citep{Zhao2020} handle parameter drift. Recent work gives variance-dependent regret bounds \citep{Wang2024}, reward dependence on recent actions \citep{Clerici2024}, latent action-dependent dynamics \citep{Choi2025}, and fixed-budget identification \citep{Xiong2024}. Our setting instead uses task-profile contexts and estimated vector-valued surrogate progress, motivating our vector-valued restarted selector SPG-Select-R for this surrogate model.

\section{Algorithms}
\label{app:algorithms}

In this paper, we use \(t\in{1,\ldots,T}\) to index bandit steps and \(i\in{1,\ldots,N}\) to index warmup states. With a slight abuse of notation, we write \(\widehat{s}_i\) during the warmup phase and \(\widehat{s}_t\) during the bandit phase. Algorithm~\ref{alg:spg} specifies the end-to-end SPG-Select workflow. Its inputs separate quantities fixed before online interaction, including the task pool, warmup budget, feature dimension, and MIRT skill dimension, from the online-control parameters \(H\), \(\gamma_g\), and \(\lambda\), and the rule that computes \(\beta_t\) from the current design matrix or sets it to a fixed experimental value. In the warmup phase, balanced randomized task selection produces calibration outcomes. These outcomes are used to fit the MIRT state-space model and construct surrogate-progress labels, each equal to a temporal difference between consecutive profile estimates, for the normalized encoder. The item parameters and encoder are then fixed, and the algorithm initializes the predictive density, its mean, and an empty regularized ridge head. 

In the bandit phase, Algorithm~\ref{alg:spg} implements the vector-valued SPG-Select-R policy described in Section~\ref{sec:method}. It first checks whether a new epoch has begun. If so, it resets only \(\mathbf{A}_t\), \(\mathbf{B}_t\), and \(\widehat{\mathbf{W}}_t\). The MIRT item parameters, filtering state, encoder, and backend-specific skill-evolution state persist across epochs. For every task \(\tau\in\mathcal{T}\), it constructs a task-profile feature, predicts vector-valued surrogate progress, and evaluates the gap-weighted UCB score. The selected maximizing task is executed. Algorithm~\ref{alg:bayes-update} supplies the next predictive density and its mean from the observed outcome. By definition, the temporal difference between consecutive predictive means is the surrogate-progress vector, which updates the current epoch's ridge statistics. The execution trajectory is then used for the backend memory update. These quantities constitute the maintained selector state. The task pool stays fixed as the skill-evolution state and profile estimates change across rounds.

Algorithm~\ref{alg:bayes-update} gives the filtering operation used after each selected interaction. It first reweights the pre-observation density \(b_t\) by the selected item's MIRT likelihood and normalizes the resulting posterior \(b_t^+\). It then propagates this posterior through the state-transition model to obtain the pre-decision density \(b_{t+1}\) for the next round. The returned predictive mean \(\widehat{\boldsymbol{s}}_{t+1}\), which is the next profile estimate, is consequently based only on observations through round \(t\). This preserves the nonanticipating ordering required by the selector. In practice, the same constrained Laplace or variational approximation used for the MIRT state-space model can implement the normalization, propagation, and predictive-mean calculation.

\begin{figure}[t]
\centering
\begin{minipage}{\linewidth}
\algorithmtoprule
\captionof{algorithm}{SPG-Select-R}
\label{alg:spg}
\algorithmmidrule
\begin{algorithmic}
\STATE Input. Task pool \(\mathcal{T}\), warmup length \(N\in\mathbb{N}_+\), bandit horizon \(T\in\mathbb{N}_+\), restart period \(H\in\{1,\ldots,T\}\), feature dimension \(d_f\in\mathbb{N}_+\), gap temperature \(\gamma_g>0\), confidence-radius rule returning \(\beta_t\geq0\), regularizer \(\lambda>0\), skill dimension \(K\in\mathbb{N}_+\)
\STATE Phase 1. Warmup
\FOR{\(i = 1\) to \(N\)}
    \STATE Select \(\tau_i\) by balanced randomized sampling
    \STATE Execute \(\tau_i\), observe \(y_i\in\{0,1\}\), and update the backend memory
    \STATE Store \((\tau_i,y_i)\)
\ENDFOR
\STATE Fit the MIRT state-space model on \(\{(\tau_i,y_i)\}_{i=1}^{N}\); fit an embedding-to-item-parameter ridge regression and predict parameters for every task in the task pool used for evolution; freeze \(\{\widehat{\boldsymbol{a}}_\tau,\widehat d_\tau\}_{\tau\in\mathcal{T}}\) and construct surrogate-progress temporal-difference labels
\STATE Train and freeze \(\widehat\Phi\) on the warmup data
\STATE Initialize \(b_1\) and the profile \(\widehat{\boldsymbol{s}}_1\) from the fitted warmup marginal of \(\boldsymbol{s}_{N+1}\), reset the round index, and set \(\mathbf{A}_1=\lambda\mathbf{I}_{d_f}\), \(\mathbf{B}_1=\mathbf{0}\), and \(\widehat{\mathbf{W}}_1=\mathbf{0}\)
\STATE Phase 2. Bandit
\FOR{\(t = 1\) to \(T\)}
    \IF{\(t>1\) and \((t-1)\bmod H=0\)}
        \STATE \(\mathbf{A}_t\gets\lambda\mathbf{I}_{d_f}\), \(\mathbf{B}_t\gets\mathbf{0}\), and \(\widehat{\mathbf{W}}_t\gets\mathbf{0}\)
    \ENDIF
    \STATE Compute \(\widehat{\boldsymbol{g}}_t=\operatorname{softmax}((\mathbf{1}-\widehat{\boldsymbol{s}}_t)/\gamma_g)\)
    \FOR{each \(\tau \in \mathcal{T}\)}
        \STATE \(\boldsymbol{\phi}_{t,\tau}=\widehat\Phi(\operatorname{cat}(\boldsymbol{e}_\tau,\widehat{\boldsymbol{s}}_t))\), \(\widetilde{\boldsymbol{\delta}}_t(\tau)=\widehat{\mathbf{W}}_t^\top\boldsymbol{\phi}_{t,\tau}\)
        \STATE \(\mathcal{U}_t(\tau)\gets\widehat{\boldsymbol{g}}_t^\top\widetilde{\boldsymbol{\delta}}_t(\tau)+\beta_t\lVert\widehat{\boldsymbol{g}}_t\rVert_2\sqrt{\boldsymbol{\phi}_{t,\tau}^\top\mathbf{A}_t^{-1}\boldsymbol{\phi}_{t,\tau}}\)
    \ENDFOR
    \STATE Select \(\tau_t\in\arg\max_{\tau\in\mathcal{T}}\mathcal{U}_t(\tau)\), breaking ties deterministically
    \STATE Execute \(\tau_t\) and observe \(y_t\) and the execution trajectory
    \STATE \((b_{t+1},\widehat{\boldsymbol{s}}_{t+1})\gets\textsc{BayesUpdate}(b_t,\widehat{\boldsymbol{a}}_{\tau_t},\widehat d_{\tau_t},y_t)\)
    \STATE \(\widehat{\boldsymbol{\delta}}_t\gets\widehat{\boldsymbol{s}}_{t+1}-\widehat{\boldsymbol{s}}_t\) and \(\boldsymbol{\phi}_t\gets\boldsymbol{\phi}_{t,\tau_t}\)
    \STATE \(\mathbf{A}_{t+1}\gets\mathbf{A}_t+\boldsymbol{\phi}_t\boldsymbol{\phi}_t^\top\)
    \STATE \(\mathbf{B}_{t+1}\gets\mathbf{B}_t+\boldsymbol{\phi}_t\widehat{\boldsymbol{\delta}}_t^\top\)
    \STATE \(\widehat{\mathbf{W}}_{t+1}\gets\mathbf{A}_{t+1}^{-1}\mathbf{B}_{t+1}\)
    \STATE Update the backend memory using the execution trajectory
\ENDFOR
\end{algorithmic}
\algorithmbottomrule
\end{minipage}
\begin{minipage}{\linewidth}
\algorithmtoprule
\captionof{algorithm}{BayesUpdate}
\label{alg:bayes-update}
\algorithmmidrule
\begin{algorithmic}
\STATE Input. \(b_t\), \((\widehat{\boldsymbol{a}}_{\tau_t},\widehat d_{\tau_t})\), and \(y_t\)
\STATE \(b_t^+(\boldsymbol{s})\gets p(y_t\mid\boldsymbol{s},\widehat{\boldsymbol{a}}_{\tau_t},\widehat d_{\tau_t})b_t(\boldsymbol{s})\)
\STATE Normalize \(b_t^+\) over \([0,1]^K\)
\STATE \(b_{t+1}(\boldsymbol{s}')\gets\int_{[0,1]^K}p(\boldsymbol{s}'\mid\boldsymbol{s})b_t^+(\boldsymbol{s})\,d\boldsymbol{s}\)
\STATE \(\widehat{\boldsymbol{s}}_{t+1}\gets\mathbb{E}_{b_{t+1}}[\boldsymbol{s}']\)
\STATE Output. \(b_{t+1}\) and \(\widehat{\boldsymbol{s}}_{t+1}\)
\end{algorithmic}
\algorithmbottomrule
\end{minipage}
\end{figure}

\subsection{Sliding-Window Comparator}
\label{app:sliding-window}

The sliding-window comparator SPG-Select-W uses the same frozen encoder, gap weights, and UCB score as SPG-Select-R. Following sliding-window UCB \citep{Cheung2019}, it replaces the restart index set \(\mathcal{E}_t\) in the ridge sums of Section~\ref{sec:method} by \(\mathcal{E}_t^{\mathrm W}=\{i\in\{1,\ldots,t-1\}:i\ge t-H_{\mathrm W}\}\) for a fixed window length \(H_{\mathrm W}\in\{1,\ldots,T\}\). After the window fills, each new observation displaces the oldest one, whereas SPG-Select-R clears the ridge statistics at epoch boundaries. Both policies retain the MIRT filtering state, frozen encoder, and backend knowledge.

\section{Implementation and Experimental Details}
\label{app:implementation-details}

\subsection{MIRT State-Space Estimation}
\label{app:mirt-estimation}

This subsection gives the rationale and warmup fitting details for the state-space MIRT model in Section~\ref{sec:method}.

\noindent\textbf{Why MIRT matches task selection.}
The selector observes a binary environmental outcome \(y_t\in\{0,1\}\) after each chosen task. This is precisely the observation type modeled by the Bernoulli response law
\[
    \Pr(y_t=1\mid\boldsymbol{s},\boldsymbol{a}_{\tau_t},d_{\tau_t})
    =\operatorname{sigmoid}\!\left(\boldsymbol{a}_{\tau_t}^\top\boldsymbol{s}-d_{\tau_t}\right).
\]
Here, \(\boldsymbol{s}\in[0,1]^K\) is the agent's multidimensional latent state, \(\boldsymbol{a}_{\tau_t}\in\mathbb{R}^K\) specifies how the selected task loads on its coordinates, and \(d_{\tau_t}\in\mathbb{R}\) is the task difficulty. A single failure does not by itself distinguish a difficult task from a low discrimination-weighted latent state. The model represents these two effects with separate parameters, which are fitted jointly using warmup outcomes under the identification constraints described below. A low success rate alone is therefore not treated as evidence that any particular profile coordinate is weak.

This separation is needed for allocation. A scalar competence estimate could summarize whether the agent is performing poorly overall, but it would not identify which capability currently has the larger estimated gap. MIRT instead yields \(\widehat{\boldsymbol{s}}_t\in[0,1]^K\), so the selector can form the dimension-wise weights \(\widehat{\boldsymbol{g}}_t=\operatorname{softmax}((\mathbf{1}-\widehat{\boldsymbol{s}}_t)/\gamma_g)\) defined in Section~\ref{sec:problem}. Thus, the latent profile is the state that connects binary feedback to the profile-conditioned allocation rule, rather than an interchangeable scalar summary.

\noindent\textbf{State dynamics and online filtering.}
For generic item parameters \((\boldsymbol{a}_\tau,d_\tau)\), write \(p_\tau(\boldsymbol{s})=\operatorname{sigmoid}(\boldsymbol{a}_\tau^\top\boldsymbol{s}-d_\tau)\); their fitted values are denoted by \((\widehat{\boldsymbol{a}}_\tau,\widehat d_\tau)\). Each discrimination coefficient determines how one latent-state coordinate enters a task's predicted success probability. Because \(\widehat{\boldsymbol{a}}_\tau\) is constrained to be elementwise nonnegative, the fitted response \(\operatorname{sigmoid}(\widehat{\boldsymbol{a}}_\tau^\top\boldsymbol{s}-\widehat d_\tau)\) is nondecreasing in every latent-state coordinate. Any nonnegative task-pool average of the fitted responses is therefore also nondecreasing coordinatewise. Anchor items or equivalent constraints fix the latent-state orientation and scale.

For latent states \(\boldsymbol{s}_i\), we use
\(
    p(\boldsymbol{s}_{i+1}\mid\boldsymbol{s}_i)=\mathcal{N}_{[0,1]^K}(\boldsymbol{s}_i,\boldsymbol{\Sigma}_s),
\)
where \(\boldsymbol{\Sigma}_s\in\mathbb{R}^{K\times K}\) is a fixed positive-definite covariance matrix and \(\mathcal{N}_{[0,1]^K}\) is truncated to \([0,1]^K\). This transition regularizes the latent trajectory and adjacent profile estimates independently of the selected task. After collecting \(N\in\mathbb{N}_+\) warmup outcomes \(\mathcal{D}_{N}=\{(\tau_i,y_i)\}_{i=1}^{N}\), we fit the item parameters and latent trajectory with expectation-maximization (EM) \citep{Dempster1977}. After warmup, the fitted item parameters are fixed and the MIRT belief is filtered online. Let \(b_t(\boldsymbol{s})\) be the predictive density before round \(t\), and let \(\boldsymbol{s}'\in[0,1]^K\) be the next-state variable. Since \(y_t\) is generated from \(\boldsymbol{s}\), the update first conditions on \(y_t\) and then propagates the state:
\[
b_t^+(\boldsymbol{s})
\propto p(y_t\mid\boldsymbol{s},\widehat{\boldsymbol{a}}_{\tau_t},\widehat d_{\tau_t})b_t(\boldsymbol{s}),
\qquad
b_{t+1}(\boldsymbol{s}')
=\int_{[0,1]^K}p(\boldsymbol{s}'\mid\boldsymbol{s})b_t^+(\boldsymbol{s})\,d\boldsymbol{s},
\qquad
\widehat{\boldsymbol{s}}_{t+1}=\mathbb{E}_{b_{t+1}}[\boldsymbol{s}'].
\]
Here, \(b_t^+\) is the posterior for the current latent state after observing \(y_t\), and \(b_{t+1}\) predicts the next latent state. Its mean \(\widehat{\boldsymbol{s}}_{t+1}\), the next profile estimate, is available before the next selection. Algorithm~\ref{alg:bayes-update} gives the complete filtering procedure.

The initial-state prior is centered at the vector whose coordinates all equal \(0.5\). Let \(\boldsymbol{\theta}=\{(\boldsymbol{a}_\tau,d_\tau)\}_{\tau\in\mathcal{T}}\) collect the generic item parameters. For \(y\in\{0,1\}\), the Bernoulli mass function is
\[
    p(y\mid\boldsymbol{s},\boldsymbol{a}_\tau,d_\tau)
    =p_\tau(\boldsymbol{s})^y[1-p_\tau(\boldsymbol{s})]^{1-y}.
\]
Together with the transition model above, this gives the complete-data log density
\[
\begin{aligned}
\ell_c(\boldsymbol{\theta};\boldsymbol{s}_{1:N+1},\mathcal{D}_{N})
&=\log p(\boldsymbol{s}_1)+\sum_{i=1}^{N}\log p(\boldsymbol{s}_{i+1}\mid\boldsymbol{s}_i)\\
&\quad+\sum_{i=1}^{N}\!\left[
y_i\log p_{\tau_i}(\boldsymbol{s}_i)+(1-y_i)\log\!\bigl(1-p_{\tau_i}(\boldsymbol{s}_i)\bigr)\right].
\end{aligned}
\]
The idealized fitting procedure uses EM. At iteration \(m\), the idealized E-step forms the latent-state posterior and its means
\[
\zeta^{(m)}(\boldsymbol{s}_{1:N+1})
=p(\boldsymbol{s}_{1:N+1}\mid\mathcal{D}_{N},\boldsymbol{\theta}^{(m)}),
\qquad
\widehat{\boldsymbol{s}}_i^{(m)}=\mathbb{E}_{\zeta^{(m)}}[\boldsymbol{s}_i].
\]
When the maximum is attained, the corresponding exact M-step solves
\[
\boldsymbol{\theta}^{(m+1)}
\in\arg\max_{\boldsymbol{\theta}\in\Theta}
\left\{\mathbb{E}_{\zeta^{(m)}}\!\left[\ell_c(\boldsymbol{\theta};\boldsymbol{s}_{1:N+1},\mathcal{D}_{N})\right]
-\frac{\eta}{2}\sum_{\tau\in\mathcal{T}}\lVert\boldsymbol{a}_\tau\rVert_2^2\right\},
\]
where \(\eta>0\) is the item-parameter regularization coefficient and \(\Theta\) imposes the nonnegativity and identification constraints. A constrained Laplace or variational smoother can approximate the nonconjugate E-step, and the limited-memory Broyden-Fletcher-Goldfarb-Shanno algorithm with box constraints (L-BFGS-B) \citep{Byrd1995} can approximately solve an M-step whose constraints are represented by box bounds. With an exact E-step, an approximate M-step is a generalized-EM update if it increases the penalized expected complete-data objective. An approximate E-step requires a separate analysis of the chosen approximation to establish monotonicity. 
The notation \(\widehat{\boldsymbol{s}}_i\) denotes the resulting approximate smoothed mean at warmup index \(i\). These warmup labels may use all \(N\) warmup outcomes. Online estimates use only observations available before each decision, and the fitted marginal of \(\boldsymbol{s}_{N+1}\) initializes the bandit phase.

After warmup fitting, a multi-output ridge regression maps embeddings to their fitted discrimination vectors and difficulties. It assigns predictions to every task in the task pool used for evolution, including tasks not sampled during warmup; these are projected to nonnegative values and held fixed during online selection. Thus, unobserved tasks do not require independently fitted MIRT item parameters.

\subsection{Experimental Details}
\label{app:experimental-details}

Tables~\ref{tab:common-experimental-details} and~\ref{tab:backend-hyperparameters} summarize the common experimental protocol and backend-specific configurations. Single-seed WebShop component ablations appear in Table~\ref{tab:webshop-ablations}.

\begin{table}[t]
\centering
\small
\setlength{\tabcolsep}{6pt}
\renewcommand{\arraystretch}{1.12}
\caption{Common computational resources and evaluation protocol.}
\label{tab:common-experimental-details}
\begin{tabular*}{\linewidth}{@{\extracolsep{\fill}}p{0.35\linewidth}p{0.55\linewidth}@{}}
\toprule
Item & Value \\
\midrule
Graphics processing units (GPUs) & \(2\times\) NVIDIA RTX A6000 \\
Task-embedding model & Qwen3-Embedding-0.6B; frozen \\
Rollouts / interaction & \(1\) \\
Eval. attempts / task & \(1\) \\
ALFWorld \(H_{\max}\) & \(50\) \\
WebShop \(H_{\max}\) & \(50\) \\
\bottomrule
\end{tabular*}
\end{table}

\begin{table}[t]
\centering
\footnotesize
\setlength{\tabcolsep}{4pt}
\renewcommand{\arraystretch}{1.08}
\caption{Backend-specific hyperparameters.}
\label{tab:backend-hyperparameters}
\begin{tabular*}{\linewidth}{@{\extracolsep{\fill}}p{0.17\linewidth}p{0.27\linewidth}p{0.47\linewidth}@{}}
\toprule
Backend & Component & Value \\
\midrule
\multirow{3}{*}{SkillRL}
& Response length & \(L=512\) tokens \\
& Skill retrieval & top-\(k=6\) \\
& Skill update & interval \(=5\), threshold \(\theta=0.4\), max. new skills \(=3\) \\
\midrule
\multirow{4}{*}{ExpeL}
& Embedder / retrieval & Qwen3-Embedding-0.6B; task-similarity top-\(k=2\) \\
& Decode & actor / evaluation temperature \(=0.3\); reflection / insight temperature \(=0\) \\
& Reflection & max. retries \(=3\) \\
& Environment & success examples \(n=(8,8)\) for (ALFWorld, WebShop) \\
\midrule
\multirow{2}{*}{Shared Reflexion}
& Episodic memory & per-task capacity \(C=3\) reflections \\
& Trial budget & max. trials / task \(=3\) \\
\bottomrule
\end{tabular*}
\end{table}

\begin{table}[t]
\centering
\footnotesize
\setlength{\tabcolsep}{4pt}
\renewcommand{\arraystretch}{1.08}
\caption{Language models used by the integrated backends.}
\label{tab:backend-language-models}
\begin{tabular*}{\linewidth}{@{\extracolsep{\fill}}p{0.17\linewidth}p{0.39\linewidth}p{0.32\linewidth}@{}}
\toprule
Backend & Task execution and evaluation & Skill evolution \\
\midrule
SkillRL & SkillRL-trained Qwen2.5-7B-Instruct & DeepSeek-V4-Flash \\
ExpeL & {\raggedright Qwen2.5-7B-Instruct\par} & DeepSeek-V4-Flash \\
Shared Reflexion & {\raggedright Qwen2.5-7B-Instruct\par} & DeepSeek-V4-Flash \\
\bottomrule
\end{tabular*}
\end{table}

\begin{table}[t]
\centering
\small
\setlength{\tabcolsep}{8pt}
\renewcommand{\arraystretch}{1.12}
\caption{SPG-Select hyperparameters.}
\label{tab:selector-hyperparameters}
\resizebox{\linewidth}{!}{%
\begin{tabular}{l cccccccc}
\toprule
Hyperparameter & \(K\) & \(d_f\) & \(\lambda\) & \(\beta_t\) & \(\gamma_g\) & \(H\) & \(N\) & \(T\) \\
\midrule
Value & \(4\) & \(16\) & \(1.0\) & \(0.1\) & \(0.1\) & \(50\) & \(56\) (ALFWorld) / \(400\) (WebShop) & \(84\) (ALFWorld) / \(600\) (WebShop) \\
\bottomrule
\end{tabular}
}
\end{table}

\begin{table}[t]
\centering
\footnotesize
\setlength{\tabcolsep}{4pt}
\renewcommand{\arraystretch}{1.12}
\caption{WebShop ablations (\%). Deltas vs. SPG-Select-R.}
\label{tab:webshop-ablations}
\begin{tabular*}{\linewidth}{@{\extracolsep{\fill}}llcc@{}}
\toprule
Backend & Setting & Score & Succ. \\
\midrule
\multirow{4}{*}{SkillRL}
& Uniform & 59.4{\scriptsize\,(-5.5)} & 33.8{\scriptsize\,(-4.0)} \\
& w/o UCB & 62.0{\scriptsize\,(-2.9)} & 32.8{\scriptsize\,(-5.0)} \\
& w/o gap & \underline{64.3}{\scriptsize\,(-0.6)} & \underline{35.6}{\scriptsize\,(-2.2)} \\
& SPG-Select-R & \textbf{64.9} & \textbf{37.8} \\
\midrule
\multirow{4}{*}{ExpeL}
& Uniform & 51.8{\scriptsize\,(+4.9)} & \underline{21.0}{\scriptsize\,(-1.6)} \\
& w/o UCB & \underline{52.2}{\scriptsize\,(+5.3)} & 20.2{\scriptsize\,(-2.4)} \\
& w/o gap & \textbf{54.8}{\scriptsize\,(+7.8)} & 19.4{\scriptsize\,(-3.2)} \\
& SPG-Select-R & 46.9 & \textbf{22.6} \\
\midrule
\multirow{4}{*}{Shared Reflexion}
& Uniform & 23.7{\scriptsize\,(-22.3)} & 11.2{\scriptsize\,(-10.0)} \\
& w/o UCB & \underline{45.1}{\scriptsize\,(-0.9)} & 17.6{\scriptsize\,(-3.6)} \\
& w/o gap & 44.9{\scriptsize\,(-1.1)} & \underline{20.0}{\scriptsize\,(-1.2)} \\
& SPG-Select-R & \textbf{46.0} & \textbf{21.2} \\
\bottomrule
\end{tabular*}
\end{table}

\FloatBarrier

Table~\ref{tab:webshop-ablations} reports the single-seed WebShop component ablations. SPG-Select-R improves terminal success over w/o UCB by 5.0, 2.4, and 3.6 percentage points for SkillRL, ExpeL, and Shared Reflexion, respectively, and over w/o gap by 2.2, 3.2, and 1.2 points. Score does not follow terminal success uniformly. Relative to w/o gap, SPG-Select-R is 0.6 points higher for SkillRL, 7.8 points lower for ExpeL, and 1.1 points higher for Shared Reflexion.

In Table~\ref{tab:common-experimental-details}, \(H_{\max}\) denotes the maximum number of environment steps. In Table~\ref{tab:backend-hyperparameters}, \(L\) denotes the response-length cap, \(k\) the retrieval count, \(\theta\) the SkillRL update threshold, \(n\) the ExpeL successful-example group size, and \(C\) Reflexion episodic-memory capacity. Parenthesized pairs follow the order (ALFWorld, WebShop).

Table~\ref{tab:backend-language-models} separates the model that acts in the environment during practice and held-out evaluation from the model that updates reusable knowledge. All SkillRL experiments use the Qwen2.5-7B-Instruct checkpoint trained by the original SkillRL pipeline for task execution, whereas the ExpeL and Shared Reflexion experiments use Qwen2.5-7B-Instruct without backend-specific training. DeepSeek-V4-Flash performs skill evolution in all three backends, including skill generation, insight updating, and reflection generation as applicable. These model choices differ from those in the original SkillRL, ExpeL, and Reflexion studies and define the integrated experimental setting evaluated here. Within each backend, the task-execution and skill-evolution models are held fixed across all acquisition policies and ablations.

Table~\ref{tab:selector-hyperparameters} reports selector settings shared by ALFWorld and WebShop. \(K\) is the number of latent profile dimensions, and \(d_f\) is the dimension of the frozen task-profile representation. \(\lambda\) regularizes the ridge head, the fixed experimental coefficient \(\beta_t=0.1\) scales its UCB confidence bonus, and \(\gamma_g\) controls how strongly the gap weights concentrate on lower estimated profile coordinates. The theoretical guarantee uses the calibrated radii in~\eqref{eq:theory-beta}. \(H\) is the number of rounds between ridge-statistic restarts. \(N\) is the warmup length and \(T\) is the bandit horizon. The total number of evolution rounds is \(N+T=140\) for ALFWorld and \(N+T=1000\) for WebShop. w/o UCB sets \(\beta_t=0\) while retaining the profile-gap weights. w/o gap retains \(\beta_t=0.1\) and replaces \(\widehat{\boldsymbol{g}}_t\) with \(\mathbf{1}/K\).

\subsubsection{Surrogate Reward and Practice Success-Rate Changes}
\label{app:surrogate-practice-association}

\begin{figure}[htbp]
    \centering
    \includegraphics[width=0.84\linewidth]{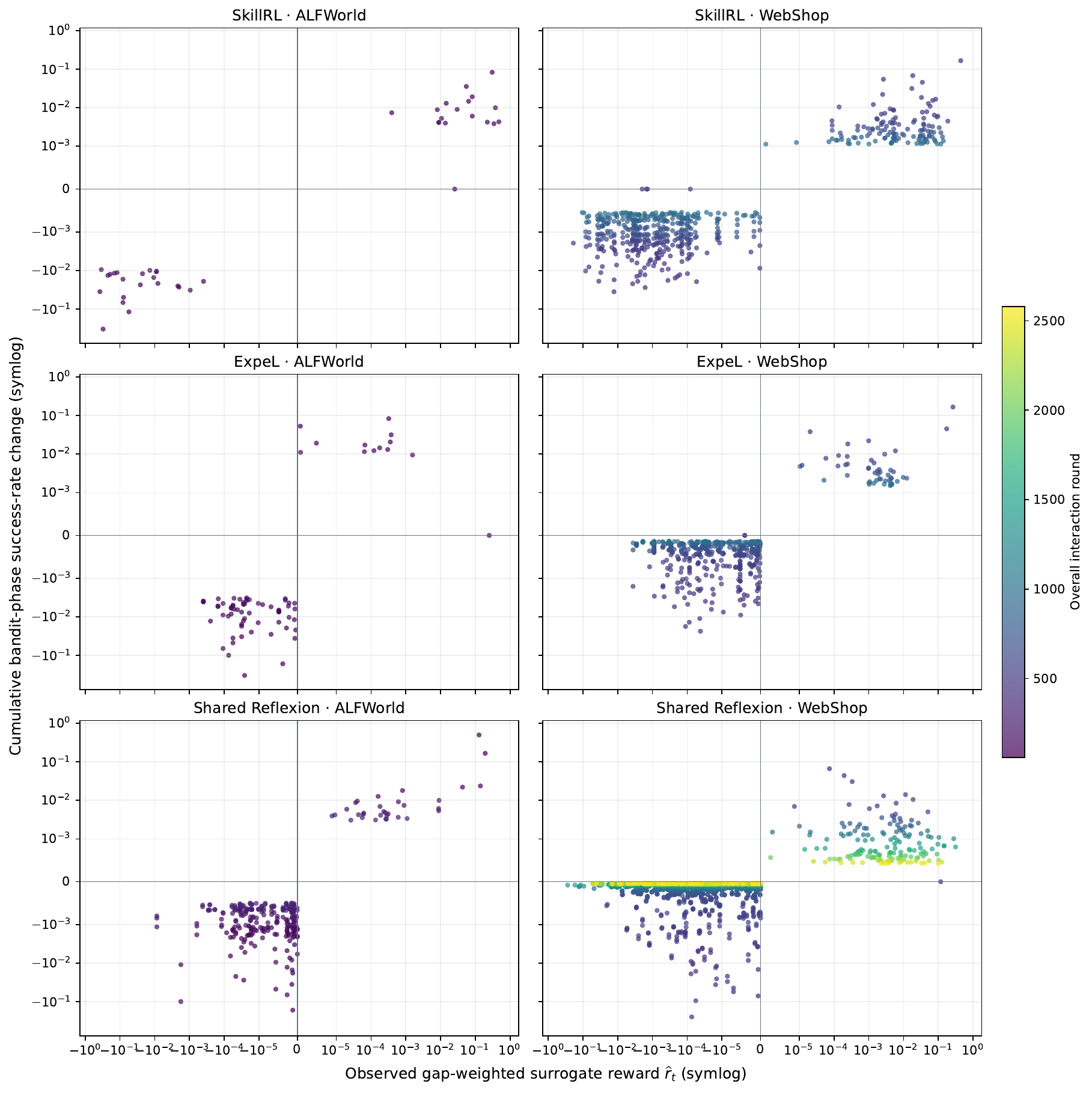}
    \caption{Observed gap-weighted surrogate reward versus the change in cumulative bandit-phase practice success rate. Rows show SkillRL, ExpeL, and Shared Reflexion, and columns show ALFWorld and WebShop.}
    \label{fig:reward-practice-association}
\end{figure}

For the single-seed reference runs, Figure~\ref{fig:reward-practice-association} relates the observed gap-weighted surrogate reward \(\widehat r_t\) in~\eqref{eq:gap-weights} to the contemporaneous change in cumulative practice success rate. Let \(y_t\in\{0,1\}\) be the outcome of bandit-phase round \(t\), and define \(C_t=t^{-1}\sum_{i=1}^t y_i\), excluding warmup. The plotted vertical value is
\[
\Delta C_t=C_t-C_{t-1}=\frac{1}{t}(y_t-C_{t-1}).
\]
We plot rounds with both a preceding bandit-phase cumulative rate and a subsequent logged pre-decision profile. This excludes the first bandit round, for which \(C_{t-1}\) is undefined, and the final logged round, for which the next profile is unavailable. We additionally omit rounds with \(\widehat r_t=0\). The horizontal value is the observed surrogate reward formed from the profile difference after the outcome, before the backend memory update. It is distinct from a predicted UCB score or a directly observed change in true skill.

The six panels share horizontal and vertical ranges. Both axes use symmetric logarithmic scales, with linear regions \([-10^{-5},10^{-5}]\) for reward and \([-0.001,0.001]\) for success-rate change. The vertical and horizontal reference lines mark zero reward and zero change, respectively. Thus, the upper-right and lower-left regions show agreement in the signs of surrogate feedback and the running-rate change, while the other two regions show disagreement. Color denotes the overall interaction round, including warmup. The figure shows the individual practice interactions, not held-out evaluation results.

Let \(\rho_{\mathrm P}\) denote the empirical Pearson correlation computed from the untransformed plotted values. For SkillRL, ExpeL, and Shared Reflexion, respectively, \(\rho_{\mathrm P}\) is \(0.47\), \(0.03\), and \(0.64\) on ALFWorld, and \(0.57\), \(0.88\), and \(0.02\) on WebShop. The associations are positive in all six runs but vary substantially, with two effectively near zero. Large positive changes in the cumulative rate occur at relatively few rounds. Because \(\Delta C_t=(y_t-C_{t-1})/t\), earlier successful rounds can yield larger changes simply through the \(1/t\) factor; these points do not establish unusually large skill updates. Since the backend memory update follows the observed outcome, these points cannot identify that update as the cause of the same round's success. Overall, the positive correlations provide a limited descriptive check on the reward's alignment with practice outcomes, not a causal test of task selection or held-out transfer.

\FloatBarrier

\begin{figure}[htbp]
    \centering
    \includegraphics[width=0.72\linewidth]{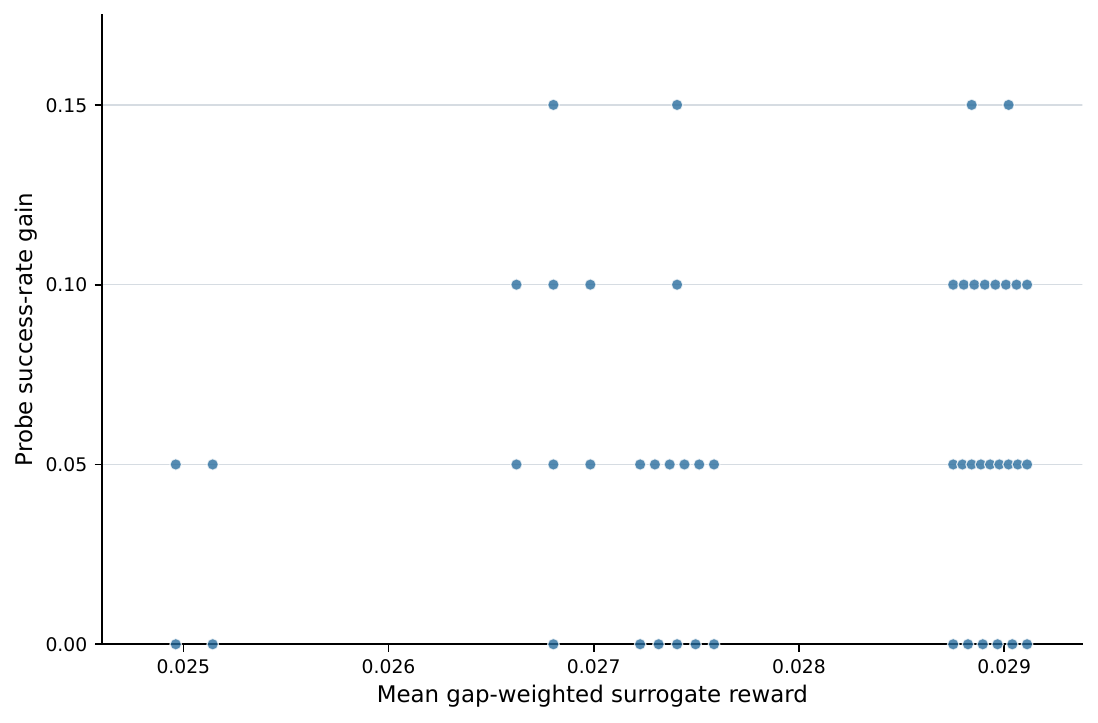}
    \caption{Mean gap-weighted surrogate reward versus probe success-rate gain for 50 ALFWorld SkillRL experiments. Each experiment evaluates the same probe tasks before and after skill evolution. Probe trajectories and outcomes are excluded from skill evolution. The displayed set contains 36 positive and 14 zero gains. A very small jitter is added to overlapping points for visual clarity.}
    \label{fig:skillrl-surrogate-probe-gain}
\end{figure}

\noindent\textbf{Probe-based transfer analysis.}
The contemporaneous analysis above cannot by itself distinguish surrogate alignment with the current outcome from alignment with transferable improvement. We therefore conduct a separate probe-based analysis. Figure~\ref{fig:skillrl-surrogate-probe-gain} shows 50 ALFWorld SkillRL experiments that compare surrogate reward with probe performance gains. Each experiment evaluates the same probe tasks before and after skill evolution. The probe uses a frozen skill state. Its trajectories and outcomes do not participate in skill evolution. All 50 displayed runs have positive mean surrogate reward. Probe success-rate gain is positive in 36 and zero in 14. These runs support using the surrogate as a reward signal for skill evolution.

Table~\ref{tab:skillrl-probe-summary} summarizes the 50 points. The variance is the sample variance across experiments.

\begin{table}[htbp]
    \centering
    \caption{Summary statistics for the 50 ALFWorld SkillRL probe-based transfer experiments.}
    \label{tab:skillrl-probe-summary}
    \begin{tabular*}{\textwidth}{@{\extracolsep{\fill}}lcc}
        \toprule
        Metric & Mean & Sample variance \\
        \midrule
        Mean gap-weighted surrogate reward \(\widehat r\) & 0.02788 & \(1.47\times10^{-6}\) \\
        Probe success-rate gain & 0.05600 & \(2.11\times10^{-3}\) \\
        \bottomrule
    \end{tabular*}
\end{table}

Each experiment applies only one skill update. Thus, the small magnitude of \(\widehat r\) is expected for a single update rather than evidence of negligible signal. Its small variance indicates that the surrogate-reward estimates are stable across the 50 experiments. The correspondingly small probe success-rate changes are consistent with the stability of the held-out probe set under a one-step update.

\FloatBarrier

\subsubsection{Backend Integration}
\label{app:backend-integration}

We retain failure-driven SkillBank expansion and skill retrieval for SkillRL \citep{Xia2026}. For ExpeL \citep{Zhao2024}, we retain trajectory-based reflection, successful-experience retrieval, and insight-bank updates. The integrated interface names its operations \textsc{Agree}, \textsc{Remove}, \textsc{Edit}, and \textsc{Add}. For Shared Reflexion, we retain Reflexion's failure-to-corrective-plan feedback mechanism \citep{Shinn2023}. The remaining changes are backend-specific. Our SkillRL integration excludes the supervised fine-tuning and reinforcement-learning stages of the complete training pipeline. ExpeL's experience gathering and insight extraction are interleaved with online task acquisition rather than executed as separate stages. Shared Reflexion extends task-local episodic memory to a cross-task retrievable memory. Consequently, the experiments evaluate SPG-Select with the skill-evolution mechanisms of SkillRL, ExpeL, and Shared Reflexion, rather than exact reproductions of their complete original pipelines.

\subsubsection{Total Evaluation API Calls}
\label{app:api-calls}

\begin{figure*}[t]
    \centering
    \includegraphics[width=\textwidth]{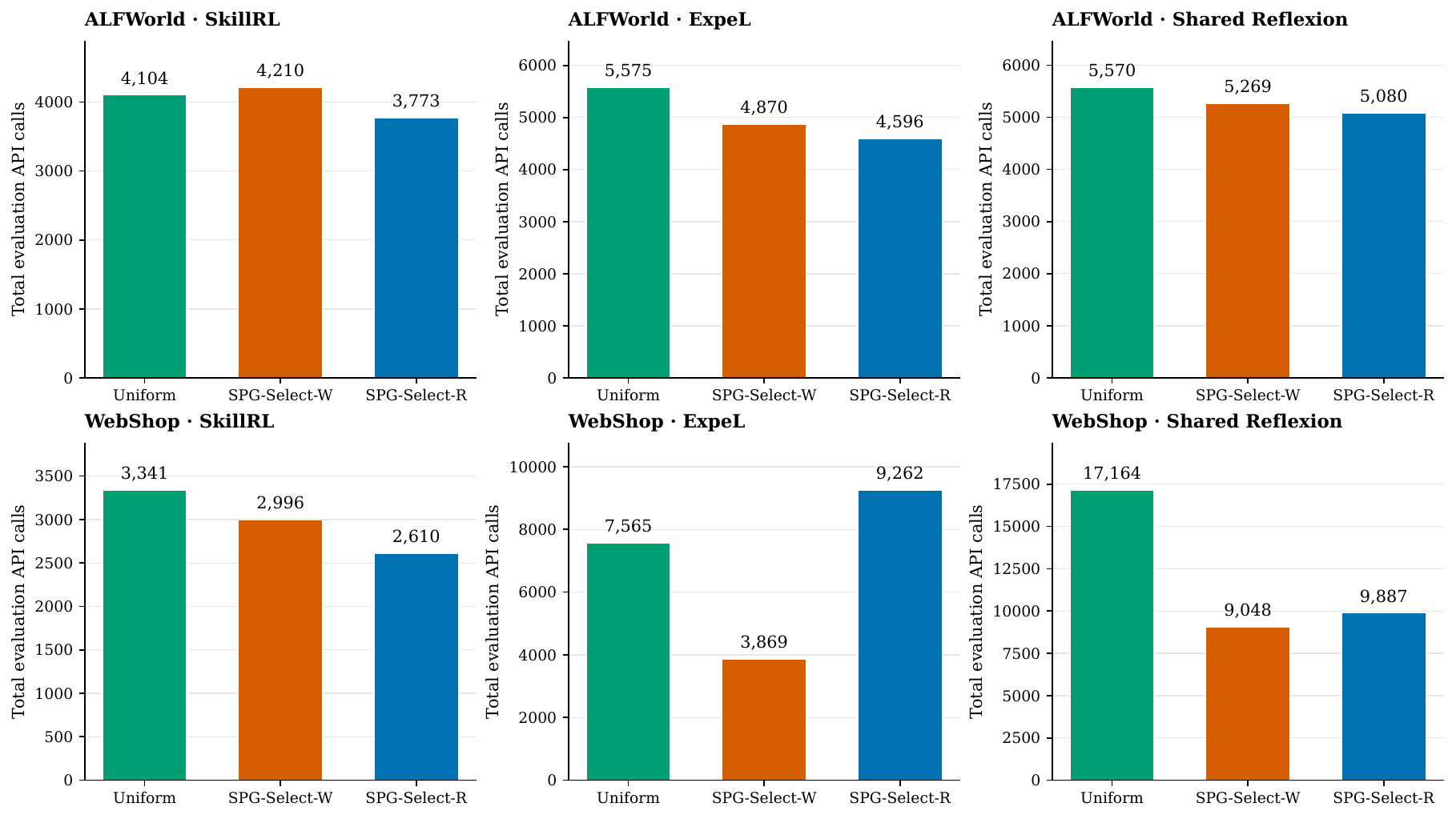}
    \caption{Total API calls over the complete held-out evaluation pool. Each bar aggregates the calls recorded for all 134 ALFWorld tasks or all 500 WebShop tasks under one task-acquisition policy. Comparisons are made within a backend and benchmark since the environments and backend prompts induce different call scales.}
    \label{fig:total-evaluation-api-calls}
\end{figure*}

Figure~\ref{fig:total-evaluation-api-calls} reports total API calls for the single-seed reference runs over the complete held-out evaluation pool. The observed total for SPG-Select-R is approximately 8\% to 42\% lower than that for Uniform in five of the six backend-benchmark combinations. The exception is ExpeL on WebShop, for which SPG-Select-R uses approximately 22\% more calls. Comparisons are made within each backend and benchmark since the environments and backend prompts induce different call scales.

\subsection{Prompt Templates and a Newly Generated Skill}
\label{app:prompts}

Following the presentation convention of the SkillRL prompt appendix \citep{Xia2026}, we report each template in a titled box. The contents below are the templates used by our vendored SkillRL adapter rather than transcriptions of SkillRL's paper. Braced terms denote values filled at runtime. The first two prompts are invoked at every environment step after retrieval, whereas Prompt~B.1 is invoked on a batch of failed trajectories to expand the skill bank.

\Needspace{19\baselineskip}
\noindent\textbf{Prompt A.1: ALFWorld Agent Execution with Retrieved Skills}

\begin{center}
\setlength{\fboxsep}{8pt}
\fbox{\begin{minipage}{0.94\linewidth}
\small
\textbf{System Prompt:}

You are an expert agent operating in the ALFRED Embodied Environment. Your task is to: \texttt{\{task\_description\}}

\textbf{\#\# Retrieved Relevant Experience}

\texttt{\{retrieved\_memories\}}

\textbf{\#\# Current Task State}

Prior to this step, you have already taken \texttt{\{step\_count\}} step(s). Below are the most recent \texttt{\{history\_length\}} observations and the corresponding actions you took: \texttt{\{action\_history\}}

You are now at step \texttt{\{current\_step\}} and your current observation is: \texttt{\{current\_observation\}}

Your admissible actions of the current situation are: [\texttt{\{admissible\_actions\}}].

Now it is your turn to take an action. First reason step-by-step about the current situation. Enclose this reasoning within \texttt{<think>} and \texttt{</think>} tags. Then choose one admissible action for the current step and present it within \texttt{<action>} and \texttt{</action>} tags.
\end{minipage}}
\end{center}

\Needspace{19\baselineskip}
\noindent\textbf{Prompt A.2: WebShop Agent Execution with Retrieved Skills}

\begin{center}
\setlength{\fboxsep}{8pt}
\fbox{\begin{minipage}{0.94\linewidth}
\small
\textbf{System Prompt:}

You are an expert autonomous agent operating in the WebShop e-commerce environment. Your task is to: \texttt{\{task\_description\}}.

\textbf{\#\# Retrieved Relevant Experience}

\texttt{\{retrieved\_memories\}}

\textbf{\#\# Current Task State}

Prior to this step, you have already taken \texttt{\{step\_count\}} step(s). Below are the most recent \texttt{\{history\_length\}} observations and the corresponding actions you took: \texttt{\{action\_history\}}

You are now at step \texttt{\{current\_step\}} and your current observation is: \texttt{\{current\_observation\}}.

Your admissible actions of the current situation are: [\texttt{\{available\_actions\}}].

Now it is your turn to take one action for the current step. First reason step-by-step about the current situation. Enclose this reasoning within \texttt{<think>} and \texttt{</think>} tags. Then choose one admissible action and present it within \texttt{<action>} and \texttt{</action>} tags. Inside \texttt{<action>}, copy exactly one native action from the admissible-action list, including its spelling and capitalization. Do not invent controls, abbreviate labels, or put the action in \texttt{<think>}.
\end{minipage}}
\end{center}

\Needspace{19\baselineskip}
\noindent\textbf{Prompt B.1: Failure Reflection and New-Skill Generation}

\begin{center}
\setlength{\fboxsep}{8pt}
\fbox{\begin{minipage}{0.94\linewidth}
\small
Analyze these failed agent trajectories and suggest NEW skills to add to the skill bank.

\textbf{FAILED TRAJECTORIES:}

\texttt{\{failure\_examples\}}

\textbf{EXISTING SKILL TITLES (avoid duplicating these):}

\texttt{\{existing\_titles\}}

Generate 1--\texttt{\{max\_new\_skills\_per\_update\}} NEW actionable skills that would help avoid these failures. Each skill must have: \texttt{skill\_id}, \texttt{title} (3--5 words), \texttt{principle} (1--2 sentences), and \texttt{when\_to\_apply}.

Use skill IDs: \texttt{\{available\_dynamic\_skill\_ids\}}.

Return ONLY a JSON array of skills, with no other text.

Example format:

\texttt{[\{"skill\_id": "dyn\_001", "title": "Verify Object Location First",}

\texttt{"principle": "Before attempting to pick up an object, verify its current location by examining the environment.",}

\texttt{"when\_to\_apply": "When a task requires moving an object but its location is uncertain"\}]}
\end{minipage}}
\end{center}

\noindent\textbf{Newly generated skill example.}
The following skill was generated for this appendix using the schema in Prompt~B.1. It is an illustrative output, separate from the skills used in the reported experimental runs. At runtime, the updater replaces the illustrative identifier with the next unused \texttt{dyn\_NNN} identifier.

\begin{center}
\small
\begin{tabular}{p{0.17\linewidth}p{0.75\linewidth}}\hline
Field & Newly generated value \\\hline
\texttt{skill\_id} & \texttt{dyn\_001} \\
\texttt{title} & Require Observable State Change \\
\texttt{principle} & After an action intended to alter the world, compare the next observation with the pre-action state and treat the transition as unconfirmed unless a goal-relevant change is visible. If no change is observed, repair an unmet precondition or choose another admissible action instead of advancing the plan. \\
\texttt{when\_to\_apply} & After an open, take, place, heat, cool, clean, or variant-selection action whose success is required by a later step. \\\hline
\end{tabular}
\end{center}

\Needspace{8\baselineskip}
\subsection{Qualitative Examples}
\label{app:qualitative-examples}

This subsection presents successful held-out evaluation trajectories from the persisted logs of the main experiments.

\subsubsection{Successful Held-Out Evaluation Cases}

The cases below show every interaction step in each successful trajectory, with model thoughts omitted. Observations are abbreviated from the persisted evaluation logs for presentation. Ellipses shorten long room inventories, option lists, or product lists without omitting any interaction step.

\begin{figure}[H]
\centering
\setlength{\fboxsep}{7pt}
\fbox{%
\begin{minipage}{0.95\linewidth}
\small
\colorbox{black!5}{\parbox{\dimexpr\linewidth-2\fboxsep\relax}{\textbf{Case 1: ALFWorld Warmed-Mug Placement (State Change and Delivery)}}}

\vspace{0.5em}
\textbf{Task:} Heat some mug and put it in the coffeemachine.

\vspace{0.35em}\hrule\vspace{0.35em}
\begin{tabular}{@{}p{0.15\linewidth}p{0.79\linewidth}@{}}
\textbf{Skills} & Systematic Exploration; Grab When Seen. \\
\textbf{Avoid} & Revisiting searched locations; ignoring visible targets. \\
\end{tabular}
\vspace{0.25em}\hrule\vspace{0.45em}

\begin{center}
\textbf{Agent Trajectory with Retrieved Skill Guidance}
\end{center}
\begin{tabular}{@{}p{0.12\linewidth}p{0.82\linewidth}@{}}
\textbf{Step 1} & \textbf{Obs:} \texttt{-= Welcome to TextWorld, ALFRED! =-} You are in the middle of a room. You see cabinets 1--6, coffeemachine 1, countertops 1--3, drawers 1--3, fridge 1, garbagecan 1, microwave 1, shelves 1--3, sinkbasin 1, stoveburners 1--4, toaster 1, \(\ldots\) \\
& \textbf{Action:} {\ttfamily go to cabinet 1} \\
\noalign{\vspace{0.2em}\hrule\vspace{0.35em}}
\textbf{Step 2} & \textbf{Obs:} You arrive at cabinet 1. On cabinet 1, you see mug 1, plate 2, and plate 1. \\
& \textbf{Action:} {\ttfamily take mug 1 from cabinet 1} \\
\noalign{\vspace{0.2em}\hrule\vspace{0.35em}}
\textbf{Step 3} & \textbf{Obs:} You pick up mug 1 from cabinet 1. \\
& \textbf{Action:} {\ttfamily go to microwave 1} \\
\noalign{\vspace{0.2em}\hrule\vspace{0.35em}}
\textbf{Step 4} & \textbf{Obs:} You arrive at microwave 1. Microwave 1 is closed. \\
& \textbf{Action:} {\ttfamily open microwave 1} \\
\noalign{\vspace{0.2em}\hrule\vspace{0.35em}}
\textbf{Step 5} & \textbf{Obs:} You open microwave 1. Microwave 1 is open; inside is egg 1. \\
& \textbf{Action:} {\ttfamily heat mug 1 with microwave 1} \\
\noalign{\vspace{0.2em}\hrule\vspace{0.35em}}
\textbf{Step 6} & \textbf{Obs:} You heat mug 1 using microwave 1. \\
& \textbf{Action:} {\ttfamily go to coffeemachine 1} \\
\noalign{\vspace{0.2em}\hrule\vspace{0.35em}}
\textbf{Step 7} & \textbf{Obs:} You arrive at coffeemachine 1. On coffeemachine 1, you see mug 2. \\
& \textbf{Action:} {\ttfamily move mug 1 to coffeemachine 1} \\
\end{tabular}

\vspace{0.35em}\hrule\vspace{0.35em}
\textbf{Result: SUCCESS} --- You move mug 1 to coffeemachine 1.
\end{minipage}}
\caption{Complete seven-step logged trajectory of successful ALFWorld evaluation task 15.}
\label{fig:alfworld-case-steps}
\end{figure}

\begin{figure}[H]
\centering
\setlength{\fboxsep}{7pt}
\fbox{%
\begin{minipage}{0.95\linewidth}
\small
\colorbox{black!5}{\parbox{\dimexpr\linewidth-2\fboxsep\relax}{\textbf{Case 2: WebShop Brown Rice Cakes (Attribute and Price Constraints)}}}

\vspace{0.5em}
\textbf{Task:} Find a 12-pack of 9-ounce certified-organic, non-GMO, gluten-free brown rice cakes priced below \$80.

\vspace{0.35em}\hrule\vspace{0.35em}
\begin{tabular}{@{}p{0.15\linewidth}p{0.79\linewidth}@{}}
\textbf{Skills} & Ground Constraints in Page Text; Click Exact Displayed Identifier. \\
\textbf{Avoid} & Unverified attribute assumptions; clicking non-controls. \\
\end{tabular}
\vspace{0.25em}\hrule\vspace{0.45em}

\begin{center}
\textbf{Agent Trajectory with Retrieved Skill Guidance}
\end{center}
\begin{tabular}{@{}p{0.12\linewidth}p{0.82\linewidth}@{}}
\textbf{Step 1} & \textbf{Obs:} ``I am looking for 9 ounce packs of 12 certified organic, non-GMO, and gluten-free organic brown rice cakes, and price lower than 80.00 dollars.'' \texttt{[Search]} \\
& \textbf{Action:} {\ttfamily search[organic brown rice cakes 9 oz 12 pack]} \\
\noalign{\vspace{0.2em}\hrule\vspace{0.35em}}
\textbf{Step 2} & \textbf{Obs:} Page 1 (50 results): \texttt{[B0013JLZBU]} Cinnamon Toast, 9.5 oz, \$100.00; \texttt{[B01GZI0JMY]} Honey Nut, 9.6 oz (pack of 12), gluten-free, USDA-certified organic, non-GMO verified, \$56.60; \texttt{[B00M8YPW0O]} Lightly Salted, 8.5 oz (pack of 12), \$68.79; \(\ldots\) \\
& \textbf{Action:} {\ttfamily click[B01GZI0JMY]} \\
\noalign{\vspace{0.2em}\hrule\vspace{0.35em}}
\textbf{Step 3} & \textbf{Obs:} Product page for Honey Nut, 9.6 oz (pack of 12), gluten-free, USDA-certified organic, non-GMO verified, \$56.60. Available sizes include \texttt{[9 ounce (pack of 12)]}; \(\ldots\) \\
& \textbf{Action:} {\ttfamily click[9 ounce (pack of 12)]} \\
\noalign{\vspace{0.2em}\hrule\vspace{0.35em}}
\textbf{Step 4} & \textbf{Obs:} You have clicked \texttt{9 ounce (pack of 12)}. The refreshed product page shows the selected size, the same certified-organic, non-GMO, gluten-free product, price \$56.60, and \texttt{[Buy Now]}. \\
& \textbf{Action:} {\ttfamily click[Buy Now]} \\
\end{tabular}

\vspace{0.35em}\hrule\vspace{0.35em}
\textbf{Result: SUCCESS} --- Environment score: 1.0.
\end{minipage}}
\caption{Complete four-step logged trajectory of successful WebShop evaluation task 46.}
\label{fig:webshop-case-steps}
\end{figure}

\FloatBarrier
\section{Proofs for Theoretical Analysis}
\label{app:proof}

This appendix states the assumptions and provides the supporting lemmas and complete proofs for Theorem~\ref{thm:dynamic-pseudoregret} and Corollary~\ref{cor:restart-selection}. Throughout, probabilities and expectations are conditional on the warmup outputs, and the assumptions below hold under this conditional law. Every epoch begins with a fresh ridge estimator, so no observation before the epoch start enters its design matrix or response statistic. The self-normalized construction follows \citet[Theorem~1]{AbbasiYadkori2011}, and we give the derivation needed for the vector-valued feedback and restarted design used here.

\subsection{Assumptions}

\begin{assumption}
\label{assump:linear-surrogate}
There are predictable conditional-mean parameter matrices \(\{\mathbf{W}_t^\star\}_{t=1}^{T}\), where \(\mathbf{W}_t^\star\in\mathbb{R}^{d_f\times K}\) is \(\mathcal{F}_{t-1}\)-measurable, such that, for every \(t\leq T\) and \(\tau\in\mathcal{T}\), \(\widehat{\boldsymbol{\delta}}_t(\tau)=\mathbf{W}_t^{\star\top}\boldsymbol{\phi}_{t,\tau}+\boldsymbol{\varepsilon}_t(\tau)\), where \(\boldsymbol{\varepsilon}_t(\tau)\in\mathbb{R}^K\) and \(\mathbb{E}[\boldsymbol{\varepsilon}_t(\tau)\mid\mathcal{F}_{t-1}]=\mathbf{0}\).
\end{assumption}

\begin{assumption}
\label{assump:boundedness}
The encoder \(\widehat\Phi\) and MIRT item-parameter estimates \(\{\widehat{\boldsymbol{a}}_\tau,\widehat d_\tau\}_{\tau\in\mathcal{T}}\) are fixed before the bandit phase, and every profile estimate used to construct a feature depends only on information available before that round. Thus, \(\widehat{\boldsymbol{s}}_t\), its gap vector \(\widehat{\boldsymbol{g}}_t\), and each feature \(\boldsymbol{\phi}_{t,\tau}\) are \(\mathcal{F}_{t-1}\)-measurable. The encoder normalization ensures \(\lVert\boldsymbol{\phi}_{t,\tau}\rVert_2\leq1\). There are positive constants \(S\) and \(\sigma\) such that, for every \(t\leq T\), \(\lVert\mathbf{W}_t^\star\rVert_F\leq S\), where \(\lVert\cdot\rVert_F\) denotes the Frobenius norm, and, for every \(\tau\in\mathcal{T}\), every \(\boldsymbol{v}\in\mathbb{R}^K\) with \(\lVert\boldsymbol{v}\rVert_2=1\), and every \(u\in\mathbb{R}\),
\[
\mathbb{E}\!\left[\exp\!\left(u\boldsymbol{v}^\top\boldsymbol{\varepsilon}_t(\tau)\right)\middle|\mathcal{F}_{t-1}\right]
\leq\exp\!\left(\frac{\sigma^2u^2}{2}\right).
\]
\end{assumption}

\begin{definition}
\label{def:path-variation}
The realized Frobenius path length of the predictable parameter sequence is \(P_T=\sum_{t=1}^{T-1}\lVert\mathbf{W}_{t+1}^\star-\mathbf{W}_t^\star\rVert_F\). This quantity may be random and may depend on the policy.
\end{definition}
By the triangle inequality and Assumption~\ref{assump:boundedness}, each increment has norm at most \(2S\), so \(0\leq P_T\leq2S(T-1)\) almost surely. The regret bound below retains \(P_T\) explicitly to quantify its dependence on parameter variation.

\subsection{Lemmas for Theorem~\ref{thm:dynamic-pseudoregret}}

Fix a restart period \(H\in\{1,\ldots,T\}\) and a failure probability \(\rho\in(0,1)\). The UCB confidence radius used in the main text is
\begin{equation}
    \beta_t=\sigma\sqrt{2K\log\!\left(
    \frac{KT\det(\mathbf{A}_t)^{1/2}}{\rho\lambda^{d_f/2}}
    \right)}+\sqrt{\lambda}S,
    \qquad
    \beta=\max_{1\leq t\leq T}\beta_t.
    \label{eq:theory-beta}
\end{equation}
Here, \(\mathbf{A}_t\) is the regularized design matrix of the current epoch. The derivation has three components. Lemma~\ref{lem:self-normalized} controls one scalar response coordinate through a self-normalized martingale inequality. Lemma~\ref{lem:epoch-confidence} applies this bound to the \(K\) coordinates of vector-valued feedback and takes a union bound over at most \(T\) restart epochs, producing the logarithmic factor in \eqref{eq:theory-beta}. The same lemma combines the coordinate-wise bounds in Frobenius norm, which yields the factor \(\sqrt{K}\), and bounds the ridge regularization term by \(\sqrt{\lambda}S\). The radii are positive because \(\sigma,S,\lambda>0\), \(KT/\rho>1\), and \(\det(\mathbf{A}_t)\geq\lambda^{d_f}\). The following lemmas provide each step in full.

Fix an epoch \(\mathcal{E}=\{t_0,\ldots,t_0+h-1\}\), where \(h\leq H\). For \(t\in\mathcal{E}\), ridge regression uses only \(i\in\{t_0,\ldots,t-1\}\). Write
\[
P(\mathcal{E}_t)=\sum_{j=t_0+1}^{t}\lVert\mathbf{W}_{j}^\star-\mathbf{W}_{j-1}^\star\rVert_F
\quad\text{and}\quad
\mathbf{A}_{\mathcal{E}}^+
=\lambda\mathbf{I}_{d_f}+\sum_{i\in\mathcal{E}}\boldsymbol{\phi}_{i}\boldsymbol{\phi}_{i}^\top.
\]

Here, \(P(\mathcal{E}_t)\) includes the change from \(\mathbf{W}_{t-1}^\star\) to the current \(\mathbf{W}_t^\star\), although \(\mathcal{E}_t\) indexes observations only through \(t-1\). All empty sums are zero, including at an epoch's first round. For any positive-definite matrix \(\mathbf{Q}\), write \(\lVert\boldsymbol{x}\rVert_{\mathbf{Q}}=\sqrt{\boldsymbol{x}^\top\mathbf{Q}\boldsymbol{x}}\).

\begin{lemma}
\label{lem:self-normalized}
Let \(\{\boldsymbol{x}_i,\xi_i\}_{i=t_0}^{t_0+h-1}\) be processes adapted to the bandit history, where \(\boldsymbol{x}_i\in\mathbb{R}^{d_f}\) is \(\mathcal{F}_{i-1}\)-measurable, \(\lVert\boldsymbol{x}_i\rVert_2\leq1\), and \(\xi_i\in\mathbb{R}\) is \(\mathcal{F}_i\)-measurable and conditionally \(\sigma\)-sub-Gaussian given \(\mathcal{F}_{i-1}\). Put \(\boldsymbol{s}_t=\sum_{i=t_0}^{t-1}\boldsymbol{x}_i\xi_i\) and \(\mathbf{V}_t=\lambda\mathbf{I}_{d_f}+\sum_{i=t_0}^{t-1}\boldsymbol{x}_i\boldsymbol{x}_i^\top\). Then, for every \(\chi\in(0,1)\), with probability at least \(1-\chi\), the following inequality holds simultaneously for all \(t\in\mathcal{E}\),
\[
\lVert\boldsymbol{s}_t\rVert_{\mathbf{V}_t^{-1}}^2
\leq 2\sigma^2\log\!\left(\frac{\det(\mathbf{V}_t)^{1/2}}{\chi\lambda^{d_f/2}}\right).
\]
\end{lemma}

\begin{proof}
Fix \(\boldsymbol{\omega}\in\mathbb{R}^{d_f}\). Conditional \(\sigma\)-sub-Gaussianity gives
\[
\mathbb{E}\!\left[\exp\!\left(\boldsymbol{\omega}^\top\boldsymbol{x}_i\xi_i-
\frac{\sigma^2}{2}(\boldsymbol{\omega}^\top\boldsymbol{x}_i)^2\right)\middle|\mathcal{F}_{i-1}\right]\leq1.
\]
Consequently,
\[
M_t(\boldsymbol{\omega})=
\exp\!\left(\boldsymbol{\omega}^\top\boldsymbol{s}_t-
\frac{\sigma^2}{2}\boldsymbol{\omega}^\top(\mathbf{V}_t-\lambda\mathbf{I}_{d_f})\boldsymbol{\omega}\right)
\]
is a nonnegative supermartingale with respect to \((\mathcal{F}_{t-1})_{t=t_0}^{t_0+h}\), with initial value one. The prefix sums and matrices are defined through \(t=t_0+h\) for this argument. Indeed, for each admissible \(t\),
\[
\begin{aligned}
\mathbb{E}[M_{t+1}(\boldsymbol{\omega})\mid\mathcal{F}_{t-1}]
&=M_t(\boldsymbol{\omega})\,
\mathbb{E}\!\left[\exp\!\left(\boldsymbol{\omega}^\top\boldsymbol{x}_t\xi_t-
\frac{\sigma^2}{2}(\boldsymbol{\omega}^\top\boldsymbol{x}_t)^2\right)\middle|\mathcal{F}_{t-1}\right]\\
&\leq M_t(\boldsymbol{\omega}),
\end{aligned}
\]
where the inequality is the preceding conditional moment-generating-function bound. Integrating with respect to the Gaussian density of \(\boldsymbol{\omega}\sim\mathcal{N}(\mathbf{0},(\sigma^2\lambda)^{-1}\mathbf{I}_{d_f})\) preserves the supermartingale property by Tonelli's theorem. Completing the square gives
\[
\begin{aligned}
\widetilde M_t
&=\left(\frac{\sigma^2\lambda}{2\pi}\right)^{d_f/2}
\int_{\mathbb{R}^{d_f}}
\exp\!\left(\boldsymbol{\omega}^\top\boldsymbol{s}_t-
\frac{\sigma^2}{2}\boldsymbol{\omega}^\top\mathbf{V}_t\boldsymbol{\omega}\right)d\boldsymbol{\omega}\\
&=\left(\frac{\sigma^2\lambda}{2\pi}\right)^{d_f/2}
\exp\!\left(\frac{\lVert\boldsymbol{s}_t\rVert_{\mathbf{V}_t^{-1}}^2}{2\sigma^2}\right)
\int_{\mathbb{R}^{d_f}}
\exp\!\left(-\frac{\sigma^2}{2}\left\lVert\boldsymbol{\omega}-\frac{\mathbf{V}_t^{-1}\boldsymbol{s}_t}{\sigma^2}\right\rVert_{\mathbf{V}_t}^2\right)d\boldsymbol{\omega}\\
&=
\frac{\lambda^{d_f/2}}{\det(\mathbf{V}_t)^{1/2}}
\exp\!\left(\frac{\lVert\boldsymbol{s}_t\rVert_{\mathbf{V}_t^{-1}}^2}{2\sigma^2}\right).
\end{aligned}
\]
The Gaussian integral is finite because \(\mathbf{V}_t\succeq\lambda\mathbf{I}_{d_f}\succ0\), and the mixture has initial value one. To obtain the maximal bound, let \(\kappa\) be the first \(t\in\mathcal{E}\) with \(\widetilde M_t\geq1/\chi\), setting \(\kappa=t_0+h\) if no crossing occurs. This is a bounded stopping time for the shifted filtration. The bounded optional stopping theorem and nonnegativity give
\[
1\geq\mathbb{E}[\widetilde M_\kappa]
\geq\chi^{-1}\Pr\!\left(\max_{t\in\mathcal{E}}\widetilde M_t\geq\chi^{-1}\right).
\]
On the complementary event, take logarithms in the displayed expression for \(\widetilde M_t\), multiply by \(2\sigma^2>0\), and rearrange to obtain the claimed inequality for every \(t\in\mathcal{E}\).
\end{proof}

\begin{lemma}
\label{lem:epoch-prefix}
Under the feature bound in Assumption~\ref{assump:boundedness}, for \(j\in\{t_0,\ldots,t-1\}\), let \(\mathbf{C}_{t,j}=\sum_{i=t_0}^{j}\boldsymbol{\phi}_i\boldsymbol{\phi}_i^\top\). Then \(\lVert\mathbf{A}_t^{-1}\mathbf{C}_{t,j}\rVert_2\leq\sqrt{d_fH/\lambda}\), where \(\lVert\cdot\rVert_2\) for a matrix denotes its operator norm induced by the Euclidean norm.
\end{lemma}

\begin{proof}
Following the matrix estimate in the proof of \citet[Lemma~2]{Zhao2020}, let \(\boldsymbol{z}\) and \(\widetilde{\boldsymbol{z}}\) be unit vectors in \(\mathbb{R}^{d_f}\) attaining the bilinear characterization of \(\lVert\mathbf{A}_t^{-1}\mathbf{C}_{t,j}\rVert_2\). By the Cauchy-Schwarz inequality in the \(\mathbf{A}_t^{-1}\)-inner product,
\[
\begin{aligned}
\left|\widetilde{\boldsymbol{z}}^\top\mathbf{A}_t^{-1}\mathbf{C}_{t,j}\boldsymbol{z}\right|
&\leq\sum_{i=t_0}^{j}\lVert\widetilde{\boldsymbol{z}}\rVert_{\mathbf{A}_t^{-1}}
\lVert\boldsymbol{\phi}_i\rVert_{\mathbf{A}_t^{-1}}
\left|\boldsymbol{\phi}_i^\top\boldsymbol{z}\right|\\
&\leq\lambda^{-1/2}\sum_{i=t_0}^{j}\lVert\boldsymbol{\phi}_i\rVert_{\mathbf{A}_t^{-1}}\\
&\leq\lambda^{-1/2}\sqrt{H\sum_{i=t_0}^{j}\lVert\boldsymbol{\phi}_i\rVert_{\mathbf{A}_t^{-1}}^2}\\
&=\lambda^{-1/2}\sqrt{H\operatorname{tr}(\mathbf{A}_t^{-1}\mathbf{C}_{t,j})}.
\end{aligned}
\]
The second inequality uses \(\mathbf{A}_t\succeq\lambda\mathbf{I}_{d_f}\), which yields \(\lVert\widetilde{\boldsymbol{z}}\rVert_{\mathbf{A}_t^{-1}}\leq\lambda^{-1/2}\), together with \(\lvert\boldsymbol{\phi}_i^\top\boldsymbol{z}\rvert\leq\lVert\boldsymbol{\phi}_i\rVert_2\lVert\boldsymbol{z}\rVert_2\leq1\). The third inequality follows from the Cauchy-Schwarz inequality and uses at most \(H\) summands. Since \(j\leq t-1\),
\[
\mathbf{C}_{t,j}\preceq\sum_{i=t_0}^{t-1}\boldsymbol{\phi}_i\boldsymbol{\phi}_i^\top
\preceq\mathbf{A}_t.
\]
Congruence by \(\mathbf{A}_t^{-1/2}\) gives \(\mathbf{A}_t^{-1/2}\mathbf{C}_{t,j}\mathbf{A}_t^{-1/2}\preceq\mathbf{I}_{d_f}\). Taking traces and using cyclicity of the trace yields \(\operatorname{tr}(\mathbf{A}_t^{-1}\mathbf{C}_{t,j})\leq d_f\). Substitution in the preceding display proves the result.
\end{proof}

\begin{lemma}
\label{lem:epoch-confidence}
Under Assumptions~\ref{assump:linear-surrogate} and~\ref{assump:boundedness}, with probability at least \(1-\rho\), simultaneously for every restart epoch, every \(t\in\mathcal{E}\), and every \(\tau\in\mathcal{T}\),
\[
\left|\widehat{\boldsymbol{g}}_t^\top(\widehat{\mathbf{W}}_t-\mathbf{W}_t^\star)^\top\boldsymbol{\phi}_{t,\tau}\right|
\leq\beta_t\lVert\widehat{\boldsymbol{g}}_t\rVert_2\lVert\boldsymbol{\phi}_{t,\tau}\rVert_{\mathbf{A}_t^{-1}}+\sqrt{\frac{d_fH}{\lambda}}P(\mathcal{E}_t).
\]
\end{lemma}

\begin{proof}
For the selected task at round \(i\), Assumption~\ref{assump:linear-surrogate} gives \(\widehat{\boldsymbol{\delta}}_i=\mathbf{W}_i^{\star\top}\boldsymbol{\phi}_i+\boldsymbol{\varepsilon}_i\), where \(\boldsymbol{\varepsilon}_i=\boldsymbol{\varepsilon}_i(\tau_i)\). Hence
\[
\mathbf{B}_t
=\sum_{i=t_0}^{t-1}\boldsymbol{\phi}_i\widehat{\boldsymbol{\delta}}_i^\top
=\sum_{i=t_0}^{t-1}\boldsymbol{\phi}_i\boldsymbol{\phi}_i^\top\mathbf{W}_i^\star
+\sum_{i=t_0}^{t-1}\boldsymbol{\phi}_i\boldsymbol{\varepsilon}_i^\top.
\]
Subtracting \(\mathbf{W}_t^\star=\mathbf{A}_t^{-1}\mathbf{A}_t\mathbf{W}_t^\star\) from \(\widehat{\mathbf{W}}_t=\mathbf{A}_t^{-1}\mathbf{B}_t\) therefore gives
\[
\widehat{\mathbf{W}}_t-\mathbf{W}_t^\star=\mathbf{A}_t^{-1}\!\left[\sum_{i=t_0}^{t-1}\boldsymbol{\phi}_i\boldsymbol{\phi}_i^\top(\mathbf{W}_i^\star-\mathbf{W}_t^\star)+\sum_{i=t_0}^{t-1}\boldsymbol{\phi}_i\boldsymbol{\varepsilon}_i^\top-\lambda\mathbf{W}_t^\star\right].
\]
For the drift term, telescope \(\mathbf{W}_i^\star-\mathbf{W}_t^\star=\sum_{j=i+1}^{t}(\mathbf{W}_{j-1}^\star-\mathbf{W}_j^\star)\). Interchanging the finite sums gives
\[
\begin{aligned}
\sum_{i=t_0}^{t-1}\boldsymbol{\phi}_i\boldsymbol{\phi}_i^\top(\mathbf{W}_i^\star-\mathbf{W}_t^\star)
&=\sum_{j=t_0+1}^{t}\left(\sum_{i=t_0}^{j-1}\boldsymbol{\phi}_i\boldsymbol{\phi}_i^\top\right)(\mathbf{W}_{j-1}^\star-\mathbf{W}_j^\star)\\
&=\sum_{j=t_0+1}^{t}\mathbf{C}_{t,j-1}(\mathbf{W}_{j-1}^\star-\mathbf{W}_j^\star).
\end{aligned}
\]
Lemma~\ref{lem:epoch-prefix}, the Frobenius-norm triangle inequality, and the definition of \(P(\mathcal{E}_t)\) now give
\[
\begin{aligned}
\left\lVert\mathbf{A}_t^{-1}\sum_{i=t_0}^{t-1}\boldsymbol{\phi}_i\boldsymbol{\phi}_i^\top(\mathbf{W}_i^\star-\mathbf{W}_t^\star)\right\rVert_F
&\leq\sum_{j=t_0+1}^{t}\lVert\mathbf{A}_t^{-1}\mathbf{C}_{t,j-1}\rVert_2\lVert\mathbf{W}_{j-1}^\star-\mathbf{W}_j^\star\rVert_F\\
&\leq\sqrt{\frac{d_fH}{\lambda}}P(\mathcal{E}_t).
\end{aligned}
\]
For the noise term, let \(\mathbf{Z}_t=\sum_{i=t_0}^{t-1}\boldsymbol{\phi}_i\boldsymbol{\varepsilon}_i^\top\). The deterministic selection rule makes \(\tau_i\) and \(\boldsymbol{\phi}_i\) \(\mathcal{F}_{i-1}\)-measurable. The observed feedback \(\widehat{\boldsymbol{\delta}}_i\) is \(\mathcal{F}_i\)-measurable, so the residual \(\boldsymbol{\varepsilon}_i=\widehat{\boldsymbol{\delta}}_i-\mathbf{W}_i^{\star\top}\boldsymbol{\phi}_i\) is also \(\mathcal{F}_i\)-measurable. Write \(\varepsilon_{i,k}\) for coordinate \(k\) of \(\boldsymbol{\varepsilon}_i\). For every coordinate \(k\), Assumption~\ref{assump:boundedness} with \(\boldsymbol{v}\) equal to the \(k\)-th standard basis vector gives conditional \(\sigma\)-sub-Gaussianity of \(\varepsilon_{i,k}(\tau)\) for every candidate \(\tau\). Using the measurability of the selected \(\tau_i\) consequently gives
\[
\mathbb{E}\!\left[\exp\!\left(u\varepsilon_{i,k}(\tau_i)\right)\middle|\mathcal{F}_{i-1}\right]
=\sum_{\tau\in\mathcal{T}}\mathbb{I}\{\tau_i=\tau\}
\mathbb{E}\!\left[\exp\!\left(u\varepsilon_{i,k}(\tau)\right)\middle|\mathcal{F}_{i-1}\right]
\leq\exp\!\left(\frac{\sigma^2u^2}{2}\right).
\]
Thus Lemma~\ref{lem:self-normalized} applies to every coordinate of \(\mathbf{Z}_t\). Apply it to coordinate \(k\) with \(\chi=\rho/(KT)\). There are at most \(T\) restart epochs and \(K\) coordinates, so a union bound gives an event of probability at least \(1-\rho\) on which, simultaneously for all epochs, times, and coordinates,
\[
\left\lVert\mathbf{A}_t^{-1/2}\sum_{i=t_0}^{t-1}\boldsymbol{\phi}_i\varepsilon_{i,k}\right\rVert_2^2
\leq2\sigma^2\log\!\left(\frac{KT\det(\mathbf{A}_t)^{1/2}}{\rho\lambda^{d_f/2}}\right).
\]
Summing over \(k=1,\ldots,K\) and taking square roots yields
\[
\lVert\mathbf{A}_t^{-1/2}\mathbf{Z}_t\rVert_F
\leq\sigma\sqrt{2K\log\!\left(\frac{KT\det(\mathbf{A}_t)^{1/2}}{\rho\lambda^{d_f/2}}\right)}.
\]
Since \(\mathbf{A}_t\succeq\lambda\mathbf{I}_{d_f}\), we have \(\mathbf{A}_t^{-1}\preceq\lambda^{-1}\mathbf{I}_{d_f}\), and hence
\[
\lVert\lambda\mathbf{A}_t^{-1/2}\mathbf{W}_t^\star\rVert_F
\leq\sqrt{\lambda}\lVert\mathbf{W}_t^\star\rVert_F
\leq\sqrt{\lambda}S.
\]
Set \(\mathbf{R}_t=\mathbf{A}_t^{-1/2}\mathbf{Z}_t-\lambda\mathbf{A}_t^{-1/2}\mathbf{W}_t^\star\). The preceding two displays and the triangle inequality give \(\lVert\mathbf{R}_t\rVert_F\leq\beta_t\). For any \(\tau\),
\[
\widehat{\boldsymbol{g}}_t^\top\left(\mathbf{A}_t^{-1}\mathbf{Z}_t-
\lambda\mathbf{A}_t^{-1}\mathbf{W}_t^\star\right)^\top\boldsymbol{\phi}_{t,\tau}
=\widehat{\boldsymbol{g}}_t^\top\mathbf{R}_t^\top\mathbf{A}_t^{-1/2}\boldsymbol{\phi}_{t,\tau}.
\]
The Cauchy-Schwarz inequality for the Frobenius inner product bounds the absolute value by \(\lVert\mathbf{R}_t\rVert_F\lVert\widehat{\boldsymbol{g}}_t\rVert_2\lVert\boldsymbol{\phi}_{t,\tau}\rVert_{\mathbf{A}_t^{-1}}\), which is at most the statistical term in the lemma. Finally, define the drift matrix
\[
\boldsymbol{g}_t=\mathbf{A}_t^{-1}\sum_{i=t_0}^{t-1}\boldsymbol{\phi}_i\boldsymbol{\phi}_i^\top(\mathbf{W}_i^\star-\mathbf{W}_t^\star).
\]
Then
\[
\left|\widehat{\boldsymbol{g}}_t^\top\boldsymbol{g}_t^\top\boldsymbol{\phi}_{t,\tau}\right|
\leq\lVert\boldsymbol{g}_t\rVert_F\lVert\widehat{\boldsymbol{g}}_t\rVert_2\lVert\boldsymbol{\phi}_{t,\tau}\rVert_2
\leq\sqrt{\frac{d_fH}{\lambda}}P(\mathcal{E}_t),
\]
where \(\lVert\widehat{\boldsymbol{g}}_t\rVert_2\leq\lVert\widehat{\boldsymbol{g}}_t\rVert_1=1\) follows from the softmax definition and \(\lVert\boldsymbol{\phi}_{t,\tau}\rVert_2\leq1\) follows from Assumption~\ref{assump:boundedness}. Combining the drift, noise, and regularization bounds proves the result.
\end{proof}

\begin{lemma}
\label{lem:epoch-width}
Under the feature bound in Assumption~\ref{assump:boundedness}, for every epoch \(\mathcal{E}\),
\[
\sum_{t\in\mathcal{E}}\lVert\boldsymbol{\phi}_{t,\tau_t}\rVert_{\mathbf{A}_t^{-1}}
\leq\sqrt{\frac{(\lambda+1)d_fh}{\lambda}\log\!\left(1+\frac{h}{\lambda d_f}\right)}.
\]
Moreover, its terminal design matrix satisfies
\[
\frac{\det(\mathbf{A}_{\mathcal{E}}^+)}{\det(\lambda\mathbf{I}_{d_f})}
\leq\left(1+\frac{h}{\lambda d_f}\right)^{d_f}.
\]
\end{lemma}

\begin{proof}
Put \(u_t=\lVert\boldsymbol{\phi}_{t,\tau_t}\rVert_{\mathbf{A}_t^{-1}}^2\). The matrix determinant lemma gives
\[
\det(\mathbf{A}_{t+1})
=\det\!\left(\mathbf{A}_t+\boldsymbol{\phi}_{t,\tau_t}\boldsymbol{\phi}_{t,\tau_t}^\top\right)
=\det(\mathbf{A}_t)(1+u_t)
\]
before any reset. Since \(\mathbf{A}_t\succeq\lambda\mathbf{I}_{d_f}\) and \(\lVert\boldsymbol{\phi}_{t,\tau_t}\rVert_2\leq1\),
\[
0\leq u_t\leq\lambda^{-1}\lVert\boldsymbol{\phi}_{t,\tau_t}\rVert_2^2\leq\frac{1}{\lambda}.
\]
For \(u\geq0\), the bound \((1+x)^{-1}\geq(1+u)^{-1}\) on \(x\in[0,u]\) gives \(\log(1+u)=\int_0^u(1+x)^{-1}\,dx\geq u/(1+u)\). Applying this with \(u=u_t\), and then using \(1+u_t\leq1+1/\lambda\), gives
\[
u_t\leq\left(1+\frac{1}{\lambda}\right)\log(1+u_t)
=\frac{\lambda+1}{\lambda}\log(1+u_t).
\]
Summing, telescoping the determinant identity, and applying the Cauchy-Schwarz inequality yield
\[
\begin{aligned}
\sum_{t\in\mathcal{E}}\sqrt{u_t}
&\leq\sqrt{h\sum_{t\in\mathcal{E}}u_t}\\
&\leq\sqrt{\frac{(\lambda+1)h}{\lambda}\sum_{t\in\mathcal{E}}\log(1+u_t)}\\
&=\sqrt{\frac{(\lambda+1)h}{\lambda}\log\!\left(\frac{\det(\mathbf{A}_{\mathcal{E}}^+)}{\det(\lambda\mathbf{I}_{d_f})}\right)}.
\end{aligned}
\]
Here, \(\mathbf{A}_{\mathcal{E}}^+\) is the terminal design matrix of the epoch, after its final observation and before any subsequent reset. To bound its determinant, let \(\nu_1,\ldots,\nu_{d_f}\geq0\) be the eigenvalues of \(\lambda^{-1}\sum_{i\in\mathcal{E}}\boldsymbol{\phi}_i\boldsymbol{\phi}_i^\top\). The arithmetic-geometric mean inequality and \(\sum_{j=1}^{d_f}\nu_j=\lambda^{-1}\sum_{i\in\mathcal{E}}\lVert\boldsymbol{\phi}_i\rVert_2^2\leq h/\lambda\) give
\[
\frac{\det(\mathbf{A}_{\mathcal{E}}^+)}{\det(\lambda\mathbf{I}_{d_f})}
=\prod_{j=1}^{d_f}(1+\nu_j)
\leq\left(1+\frac{1}{d_f}\sum_{j=1}^{d_f}\nu_j\right)^{d_f}
\leq\left(1+\frac{h}{\lambda d_f}\right)^{d_f}.
\]
Substituting this inequality into the preceding display proves the result.
\end{proof}

\subsection{Proof of Theorem~\ref{thm:dynamic-pseudoregret}}

\begin{proof}
Write \(\widehat q_t(\tau)=\widehat{\boldsymbol{g}}_t^\top\widehat{\mathbf{W}}_t^\top\boldsymbol{\phi}_{t,\tau}\), \(w_t(\tau)=\lVert\boldsymbol{\phi}_{t,\tau}\rVert_{\mathbf{A}_t^{-1}}\), and \(D_t=\sqrt{d_fH/\lambda}P(\mathcal{E}_t)\). Assumption~\ref{assump:linear-surrogate} and conditional mean-zero noise imply \(q_t(\tau)=\widehat{\boldsymbol{g}}_t^\top\mathbf{W}_t^{\star\top}\boldsymbol{\phi}_{t,\tau}\). Thus, the event in Lemma~\ref{lem:epoch-confidence} implies, for every candidate \(\tau\),
\[
q_t(\tau)\leq\widehat q_t(\tau)+\beta_t\lVert\widehat{\boldsymbol{g}}_t\rVert_2w_t(\tau)+D_t,
\qquad
q_t(\tau)\geq\widehat q_t(\tau)-\beta_t\lVert\widehat{\boldsymbol{g}}_t\rVert_2w_t(\tau)-D_t,
\]
where the first and second inequalities are respectively the upper and lower sides of that lemma. Algorithm~\ref{alg:spg} selects an action maximizing \(\widehat q_t(\tau)+\beta_t\lVert\widehat{\boldsymbol{g}}_t\rVert_2w_t(\tau)\). Applying the upper inequality to \(\tau_t^*\), UCB optimality, and then the lower inequality to \(\tau_t\), in this order, gives
\begin{align*}
q_t(\tau_t^*)-q_t(\tau_t)
&\leq \widehat q_t(\tau_t^*)+\beta_t\lVert\widehat{\boldsymbol{g}}_t\rVert_2w_t(\tau_t^*)+D_t-q_t(\tau_t)\\
&\leq \widehat q_t(\tau_t)+\beta_t\lVert\widehat{\boldsymbol{g}}_t\rVert_2w_t(\tau_t)+D_t-q_t(\tau_t)\\
&\leq2\beta_t\lVert\widehat{\boldsymbol{g}}_t\rVert_2w_t(\tau_t)+2D_t\\
&\leq2\beta_tw_t(\tau_t)+2D_t.
\end{align*}
The final inequality uses \(\lVert\widehat{\boldsymbol{g}}_t\rVert_2\leq1\). For an epoch \(\mathcal{E}=\{t_0,\ldots,t_0+h-1\}\), define \(P(\mathcal{E})=\sum_{j=t_0+1}^{t_0+h-1}\lVert\mathbf{W}_{j}^\star-\mathbf{W}_{j-1}^\star\rVert_F\). Since \(\beta_t\leq\beta\), Lemma~\ref{lem:epoch-width} and \(h\leq H\) give
\[
\sum_{t\in\mathcal{E}}2\beta_tw_t(\tau_t)
\leq2\beta\sqrt{\frac{(\lambda+1)d_fh}{\lambda}\log\!\left(1+\frac{h}{\lambda d_f}\right)}
\leq2\beta\sqrt{\frac{(\lambda+1)d_fH}{\lambda}\log\!\left(1+\frac{H}{\lambda d_f}\right)}.
\]
Moreover, expanding the definition of the local path length and interchanging the finite sums yield
\begin{align*}
\sum_{t\in\mathcal{E}}P(\mathcal{E}_t)
&=\sum_{t=t_0}^{t_0+h-1}\sum_{j=t_0+1}^{t}
\lVert\mathbf{W}_{j}^\star-\mathbf{W}_{j-1}^\star\rVert_F\\
&=\sum_{j=t_0+1}^{t_0+h-1}(t_0+h-j)
\lVert\mathbf{W}_{j}^\star-\mathbf{W}_{j-1}^\star\rVert_F\\
&\leq H\sum_{j=t_0+1}^{t_0+h-1}
\lVert\mathbf{W}_{j}^\star-\mathbf{W}_{j-1}^\star\rVert_F
=HP(\mathcal{E}),
\end{align*}
where the inequality uses \(t_0+h-j\leq h-1\leq H\). Consequently,
\[
\sum_{t\in\mathcal{E}}2D_t
=2\sqrt{\frac{d_fH}{\lambda}}\sum_{t\in\mathcal{E}}P(\mathcal{E}_t)
\leq2\sqrt{\frac{d_f}{\lambda}}H^{3/2}P(\mathcal{E}).
\]
There are exactly \(\lceil T/H\rceil\) epochs, with the last possibly shorter than \(H\). Their within-epoch variation sums satisfy \(\sum_{\mathcal{E}}P(\mathcal{E})\leq P_T\), since each increment \(\lVert\mathbf{W}_{j}^\star-\mathbf{W}_{j-1}^\star\rVert_F\) can occur in at most one within-epoch path length. Summing the preceding estimation and drift bounds over the epochs gives
\[
R_T \leq 2\beta\left\lceil\frac{T}{H}\right\rceil
\sqrt{\frac{(\lambda+1)d_fH}{\lambda}\log\!\left(1+\frac{H}{\lambda d_f}\right)}
+2\sqrt{\frac{d_f}{\lambda}}H^{3/2}P_T,
\]
which is the claimed high-probability conclusion.
\end{proof}

\subsection{Oracle Choice of the Restart Period}

For restart-period selection, define the horizon-uniform UCB radius
\begin{equation}
\beta^{\mathrm{u}}=
\sigma\sqrt{2K\left[\log\!\left(\frac{KT}{\rho}\right)
+\frac{d_f}{2}\log\!\left(1+\frac{T}{\lambda d_f}\right)\right]}
+\sqrt{\lambda}S.
\label{eq:horizon-uniform-radius}
\end{equation}

We next verify the deterministic comparison between the time-specific and horizon-uniform radii. Fix \(t\in\{1,\ldots,T\}\), and let \(\mathcal{E}=\{t_0,\ldots,t_0+h-1\}\) be the epoch containing \(t\). By the definitions of the prefix and terminal design matrices,
\[
\mathbf{A}_t
=\lambda\mathbf{I}_{d_f}+\sum_{i=t_0}^{t-1}\boldsymbol{\phi}_{i}\boldsymbol{\phi}_{i}^\top
\preceq\lambda\mathbf{I}_{d_f}+\sum_{i\in\mathcal{E}}\boldsymbol{\phi}_{i}\boldsymbol{\phi}_{i}^\top
=\mathbf{A}_{\mathcal{E}}^+.
\]
The matrix \(\mathbf{A}_t^{-1/2}\mathbf{A}_{\mathcal{E}}^+\mathbf{A}_t^{-1/2}\succeq\mathbf{I}_{d_f}\) has all eigenvalues at least one. Thus,
\[
\frac{\det(\mathbf{A}_{\mathcal{E}}^+)}{\det(\mathbf{A}_t)}
=\det\!\left(\mathbf{A}_t^{-1/2}\mathbf{A}_{\mathcal{E}}^+\mathbf{A}_t^{-1/2}\right)
\geq1.
\]
Applying this observation and Lemma~\ref{lem:epoch-width} gives
\[
\frac{\det(\mathbf{A}_t)^{1/2}}{\lambda^{d_f/2}}
\leq\frac{\det(\mathbf{A}_{\mathcal{E}}^+)^{1/2}}{\lambda^{d_f/2}}
\leq\left(1+\frac{h}{\lambda d_f}\right)^{d_f/2}
\leq\left(1+\frac{T}{\lambda d_f}\right)^{d_f/2},
\]
where the last inequality uses \(h\leq H\leq T\). Consequently,
\[
\begin{aligned}
\beta_t
&\leq\sigma\sqrt{2K\left[\log\!\left(\frac{KT}{\rho}\right)
+\frac{d_f}{2}\log\!\left(1+\frac{T}{\lambda d_f}\right)\right]}
+\sqrt{\lambda}S\\
&=\beta^{\mathrm{u}}.
\end{aligned}
\]
Since this holds for every \(t\in\{1,\ldots,T\}\), the definition \(\beta=\max_{1\leq t\leq T}\beta_t\) yields \(\beta\leq\beta^{\mathrm{u}}\).

\begin{corollary}
\label{cor:restart-selection}
Fix the warmup outputs and the parameters of Theorem~\ref{thm:dynamic-pseudoregret}. Let \(V_T\geq0\) be a deterministic budget specified before the bandit phase. Suppose that the parameter path under the policy defined below satisfies \(P_T\leq V_T\) almost surely. For \(V_T>0\), define
\[
L=(\lambda+1)\log\!\left(1+\frac{T}{\lambda d_f}\right),
\qquad
H^\circ=\left(\frac{\beta^{\mathrm{u}}T\sqrt L}{V_T}\right)^{1/2},
\qquad
H^*=\min\{T,\max\{1,\lceil H^\circ\rceil\}\}.
\]
Run Algorithm~\ref{alg:spg} with the fixed period \(H=H^*\) and the radii in \eqref{eq:theory-beta}. With probability at least \(1-\rho\), conditional on the warmup outputs,
\begin{equation}
R_T\leq\min\!\left\{T,\;
2\beta^{\mathrm{u}}\left\lceil\frac{T}{H^*}\right\rceil
\sqrt{\frac{(\lambda+1)d_fH^*}{\lambda}\log\!\left(1+\frac{H^*}{\lambda d_f}\right)}
+2\sqrt{\frac{d_f}{\lambda}}(H^*)^{3/2}V_T\right\}.
\label{eq:oracle-finite-bound}
\end{equation}
If \(1\leq H^\circ\leq T\), the same event gives
\[
R_T\leq(4+2^{5/2})\sqrt{\frac{d_f}{\lambda}}
(\beta^{\mathrm{u}}T)^{3/4}L^{3/8}V_T^{1/4}.
\]
When \(K,\lambda,\lambda^{-1},S,\sigma,\rho^{-1}\) are fixed independently of \(d_f\) and \(T\), this simplifies to
\[
R_T=\widetilde{\mathcal{O}}\!\left(d_f^{7/8}T^{3/4}V_T^{1/4}\right).
\]
For \(H^\circ>T\), the period is \(H^*=T\), and the bound is at most \(4\beta^{\mathrm{u}}\sqrt{d_fLT/\lambda}\). For \(H^\circ<1\), the period is \(H^*=1\), and \eqref{eq:oracle-finite-bound} remains valid without the interior-rate simplification. Separately, if \(V_T=0\), use \(H=T\). Then
\[
R_T\leq2\beta^{\mathrm{u}}\sqrt{\frac{(\lambda+1)d_fT}{\lambda}\log\!\left(1+\frac{T}{\lambda d_f}\right)},
\]
which is \(\widetilde{\mathcal{O}}(d_f\sqrt T)\) in the same fixed-parameter regime.
\end{corollary}

\begin{proof}
The radius \(\beta^{\mathrm{u}}\) and budget \(V_T\) are fixed before the bandit phase, so \(H^*\) is a nonanticipating integer in \(\{1,\ldots,T\}\). Apply Theorem~\ref{thm:dynamic-pseudoregret} to this period, and intersect its confidence event with the probability-one event \(P_T\leq V_T\). Substituting \(P_T\leq V_T\) and \(\beta\leq\beta^{\mathrm{u}}\) yields the second entry of the minimum in \eqref{eq:oracle-finite-bound}.

For the first entry, every candidate's potential next profile lies in \([0,1]^K\) and is compared with the same current profile \(\widehat{\boldsymbol{s}}_t\). Therefore, the coordinatewise bounds on its temporal difference and their conditional expectations give
\[
-\widehat{\boldsymbol{s}}_t\leq\boldsymbol{\mu}_t(\tau)\leq\mathbf{1}-\widehat{\boldsymbol{s}}_t.
\]
The gap weights are nonnegative and sum to one, so
\[
-\widehat{\boldsymbol{g}}_t^\top\widehat{\boldsymbol{s}}_t
\leq q_t(\tau)\leq1-\widehat{\boldsymbol{g}}_t^\top\widehat{\boldsymbol{s}}_t.
\]
This interval has length one and is common to all candidates. Hence \(0\leq q_t(\tau_t^*)-q_t(\tau_t)\leq1\) and \(R_T\leq T\) deterministically.

For \(1\leq H^*\leq T\), use \(\lceil T/H^*\rceil\leq2T/H^*\) and \(\log(1+H^*/(\lambda d_f))\leq\log(1+T/(\lambda d_f))\). The second entry of the minimum is therefore at most
\[
\sqrt{\frac{d_f}{\lambda}}\left[
\frac{4\beta^{\mathrm{u}}T\sqrt L}{\sqrt{H^*}}
+2(H^*)^{3/2}V_T\right].
\]
The unrounded choice balances the two terms without their numerical constants, since
\[
\frac{\beta^{\mathrm{u}}T\sqrt L}{\sqrt{H^\circ}}
=(\beta^{\mathrm{u}}T\sqrt L)^{3/4}V_T^{1/4}
=V_T(H^\circ)^{3/2}.
\]
If \(1\leq H^\circ\leq T\), then \(H^*=\lceil H^\circ\rceil\) and \(H^\circ\leq H^*\leq2H^\circ\). Relative to \(H^\circ\), this rounding does not increase the estimation term and increases the drift term by at most a factor of \(2^{3/2}\). The preceding bound becomes
\[
R_T\leq(4+2\cdot2^{3/2})\sqrt{\frac{d_f}{\lambda}}
(\beta^{\mathrm{u}}T\sqrt L)^{3/4}V_T^{1/4},
\]
which proves the stated constant and parameter dependence. Under the fixed-parameter regime in the corollary, \eqref{eq:horizon-uniform-radius} gives
\[
\begin{aligned}
\beta^{\mathrm{u}}
&=\sigma\sqrt{2K\left[\log\!\left(\frac{KT}{\rho}\right)
+\frac{d_f}{2}\log\!\left(1+\frac{T}{\lambda d_f}\right)\right]}
+\sqrt{\lambda}S
=\widetilde{\mathcal{O}}(\sqrt{d_f}),
\end{aligned}
\]
while \(L\) is logarithmic in \(d_f,T\). Thus the dimension factor is \(d_f^{1/2}(d_f^{1/2})^{3/4}=d_f^{7/8}\), giving the interior rate.

If \(H^\circ>T\), its definition implies \(V_T<\beta^{\mathrm{u}}\sqrt L/T\). In \eqref{eq:oracle-finite-bound}, set \(H^*=T\) to obtain
\[
\begin{aligned}
R_T
&\leq2\beta^{\mathrm{u}}\sqrt{\frac{d_fLT}{\lambda}}
+2\sqrt{\frac{d_f}{\lambda}}T^{3/2}V_T\\
&\leq4\beta^{\mathrm{u}}\sqrt{\frac{d_fLT}{\lambda}}.
\end{aligned}
\]
If \(H^\circ<1\), clipping sets \(H^*=1\), already covered by the finite-sample bound. Finally, \(V_T=0\) implies \(P_T=0\) almost surely. Substituting \(H=T\) into Theorem~\ref{thm:dynamic-pseudoregret} and using \(\beta\leq\beta^{\mathrm{u}}\) proves the stationary bound. Its factors have orders \(\widetilde{\mathcal{O}}(\sqrt{d_f})\) and \(\widetilde{\mathcal{O}}(\sqrt{d_fT})\), respectively, so their product is \(\widetilde{\mathcal{O}}(d_f\sqrt T)\).
\end{proof}

The minimum in \eqref{eq:oracle-finite-bound} makes explicit when the analysis improves on the trivial bound. For fixed \(d_f,K,\lambda,S,\sigma,\rho\), the interior expression is \(\mathcal{O}(T^{3/4}V_T^{1/4}(\log T)^{3/4})\). Thus \(V_T(\log T)^3=o(T)\) is sufficient for a sublinear bound in that regime. A finite path length alone does not guarantee sublinear regret.

\end{document}